\documentclass[a4paper,english]{article}

\pdfoutput=1
\usepackage{amsmath, calc}
\usepackage{amsfonts}

\usepackage{enumitem}
\setlist{nosep} 

\usepackage{multirow,tabularx, booktabs, colortbl} 
\usepackage{xcolor}
\usepackage{graphicx}
\usepackage[justification=centering, 
font=small, 
labelfont=bf, 
textfont=it]{caption} 

\usepackage{nameref}
\usepackage[hidelinks]{hyperref} 
\hypersetup{
    colorlinks,
    linktoc=all,
    linkcolor={darkgray},
    citecolor={darkgray},
    urlcolor={darkgray}
}

\usepackage[authordate,strict,backend=biber,babel=other,bibencoding=inputenc]{biblatex-chicago}

\newcommand{\citein}[1]{\citeauthor{#1} \autocite*{#1}}

\title{Augmented Computational Design}
\usepackage{authblk} 

\author[*1]{Pirouz Nourian}
\author[2]{Shervin Azadi}
\author[3]{Roy Uijtendaal}
\author[4]{Nan Bai}
\affil[1]{University of Twente, 7522 NH, Hallenweg 8, Enschede, NL}
\affil[2]{Eindhoven University of Technology, 5612 AZ, Het Kranenveld 8, Eindhoven, NL}
\affil[3]{Nieman Raadgevende Ingenieurs, 3542 AB, Atoomweg 400, Utrecht, NL}
\affil[4]{Delft University of Technology, 2628 BL, Julianalaan 134, Delft, NL}
\affil[*]{Corresponding Author: p.nourian@utwente.nl}
\begin{document}

\maketitle

\begin{abstract}
    This chapter presents methodological reflections on the necessity and utility of artificial intelligence in generative design.
    Specifically, the chapter discusses how generative design processes can be augmented by AI to deliver in terms of a few outcomes of interest or performance indicators while dealing with hundreds or thousands of small decisions.
    The core of the performance-based generative design paradigm is about making statistical or simulation-driven associations between these choices and consequences for mapping and navigating such a complex decision space.
    This chapter will discuss promising directions in Artificial Intelligence for augmenting decision-making processes in architectural design for mapping and navigating complex design spaces.
\end{abstract}

\footnote{This is the author version of the book chapter “Augmented Computational Design: Methodical Application of Artificial Intelligence in Generative Design.” In Artificial Intelligence in Performance-Driven Design: Theories, Methods, and Tools Towards Sustainability, edited by Narjes Abbasabadi and Mehdi Ashayeri. Wiley, 2023} 

\section{Introduction}
The core of the performance-driven computational design is to trace the sensitivity of variations of some performance indicators to the differences between design alternatives. 
Therefore any argument about the utility of AI for performance-based design must necessarily discuss the representation of such differences, as explicitly as possible. 
The existing data models and data representations in the field of Architecture, Engineering, and Construction (AEC), such as CAD and BIM are heavily focused on geometrically representing building elements and facilitating the process of construction management.
Unfortunately, the field of AEC does not currently have a structured discourse based on an explicit representation of decision variables and outcomes of interest. 
Specifically, the notion of design representation and the idea of data modelling for representing ``what needs to be attained from buildings'' is rather absent in the literature.

This treatise proposes to systematically view the differences between design alternatives in terms of decision variables, be they spatial or non-spatial. 
Based on such an explicit formulation of decision variables, we set forth a framework for building and utilizing AI in [architectural] generative design processes for associating decision variables and outcomes of interest as performance indicators in a reciprocal relationship. 
This reciprocity is explained in terms of the duality between two quintessential problems to be addressed in generative design, i.e. the problem of evaluation of design alternatives (mapping), and the problem of derivation of design alternatives (navigation). 

Starting from an explicit representation of a design space as an ordered pair of two vectors respectively denoting decision variables and performance indicators, we put forth a mathematical framework for structuring data-driven approaches to generative design in the field of AEC. 
This framework highlights two major types of applications for AI in performance-driven design and their fusion: those capable of augmenting design evaluation procedures and those capable of augmenting design derivation procedures. 
Moreover, we introduce the reciprocity of ``flows'' and ``manifolds'' as an intermediary notion for going beyond the so-called form-function dichotomy. 
Discussing these notions necessitates the introduction of a mathematical foundation for the framework rooted in multi-variate calculus. 

The main advantage of this explicit formulation is to enhance the explainability of AI when utilized in generative design by introducing meaningful and interpretable latent spaces based on the reciprocal relationship between manifolds and flows.
The balance of predictive/deterministic power and interpretability/explainability is discussed in the concrete context of an illustrative example.

A chain of key concepts will be introduced in this chapter, starting from the notion of decision-making in design, the nature of design variables, the specifics of spatial decision variables, the notion of design space, and the two dual actions in the exploration of design spaces: mapping and navigating. 

Whilst the introduced framework is quite general, a particular class of Probabilistic Graphical Models (PGM), Bayesian Belief Networks (BBN), is introduced to provide a concrete illustrative example of the utility of AI in AEC. 
For a deeper insight into this particular approach to data-driven design, the readers are referred to the two classical books about PGM: \citein{pearl_probabilistic_1988} \& \citein{koller_probabilistic_2009}. 
The illustrative example is a BBN trained for making a data-driven replica of the building energy model used by the Dutch government in order to obtain a rough meta-model to be used in mass-scale policy analysis, e.g. for advising the government on the relative utility of energy transition subsidies and planning measures. 
This example is chosen not because the BBNs are the most advanced models or the most accurate models for approximating such large functions. 
The choice is rather pragmatic in that this model has proven to be promising from the stance of predictive power while retaining a basic level of theoretical interpretability and intuitive appeal. 

The chapter is structured as follows: we first present a historical context to establish the necessity of such a data-driven generative design framework; continue with conceptualizing and mathematically formulating the structure of the framework, dubbed as Augmented Computational Design (ACD); present an illustrative example demonstrating the utility of the framework; and conclude with a discussion on its outlook, open questions, and avenues for further research. 

\section{Background}
\label{sec:background}

Here we revisit the utility of AI for data-driven generative design by highlighting some key gaps of knowledge in the field of AEC and briefly mentioning overarching frameworks in computational design and AI that can be used to address these gaps.

\subsection{Relevance of AI in AEC}
The earliest attempts for enhancing accountability and predictive power in computational design can be traced back to the notions of Scientific Architecture \autocite{friedman_toward_1975} and the Sciences of the Artificial \autocite{simon2019sciences}. 
Both of these seminal books explicitly discuss the necessity of forming some kind of a specific spatial and configurative form of design knowledge, the core of which boils down to being able to explicitly represent the main subject matter of spatial design as ``spatial configurations''. 
One of the first phenomenological and systemic descriptions of design processes explicitly referring to the notion of performance is the ``Function, Behaviour, Structure'' framework of \citein{gero2004situated}, in which the overused notions of form and function are elaborated in terms of expected and required behaviour/functionality from a system (dubbed as the function), its design as a form or configuration (dubbed as structure), and its performance (dubbed as behaviour). 
The idea of design as a process of generating the representation of a spatial structure is explicitly discussed in this framework and the difference between desired behaviour and the actual behaviour of the structure is discussed as the performance drive for the process. 
What can be observed in this phenomenological framework, predating most recent advancements in computational design, is the fundamental belief about the innate necessity of creativity in terms of the cognitive capability of designers for proposing structures capable of working as desired, based on some kind of tacit knowledge. 
Congruently, an anthropological description of design processes refers to the old duality between form (structure) and function (purpose) of designed artefacts, and the fact that [in the absence of explicit knowledge and representation schemes], as \citein{kroes2010engineering} has put it, designers are traditionally trained to produce solutions (draw them) through a ``logical leap'' often without even understanding or paying any attention to the design requirements or supposed levels of quality attainment. 
Suppose we wanted to evaluate (compare) two different alternative designs for a hospital, \autocite{jia2023spatial}, or a home, how do we want to represent the designs digitally for a computer to evaluate them? 
Let us discuss an analogical example: if we wanted to compare two pieces of music in terms of their beauty, it would be very straightforward to digitize their notations and feed them to a machine, because the musical notation is already discretized (digitized), regardless if it is written on paper or etched on the cylinder of an old-fashioned winding music-box \autocite{zeng2021musicbert}. 
However, doing the same, that is comparing two buildings, would be a much more difficult challenge especially because there is currently no [discrete/textual] notation for spatial design that can capture the features of spatial configurations.

Instead of the extensive emphasis on the product of architecture as the shapes of buildings, we turn our attention to the processes of design and put a lens of ``design as [discrete] decision-making" on the debate to avoid commonplace reduction of design to the production of design drawings. 
This view forms the basis of the generative design paradigm as extensively articulated by \citein{Nourian2023} \& \citein{veloso2021mapping}. 
Similarly, the challenges, opportunities, and promising ways of utilization of AI (particularly deep-learning and generative models) for goal-oriented design explorations have been discussed extensively in \autocite{regenwetter_deep_2022} \& \autocite{regenwetter_towards_2022}. 

\subsection{Historical Context}

In this section, we first give a very brief history of the most important and relevant developments in AI; 
then lay the foundation of a formulation of architectural design as a matter of decision-making; discuss the mathematical implications of this paradigmatic frame for generative design; elaborate on the notion of decision-making and the duality of derivation and evaluation problems; 
and discuss two statistical approaches to design, a possibilistic approach utilizing Fuzzy Logic or Markovian Design Machines and a probabilistic approach utilizing Bayesian Belief Networks or Diffusion Models. 

\begin{figure}[ht]
     \centering
     \includegraphics[width=\textwidth]{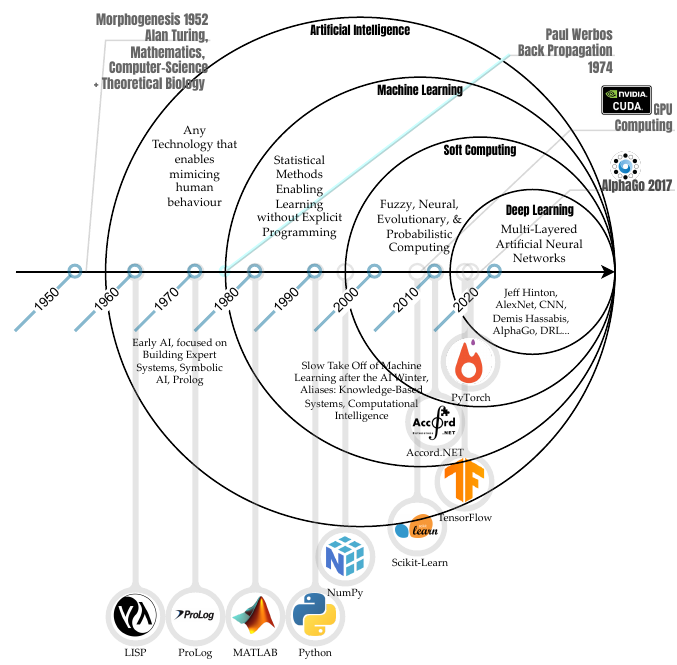}
     \caption{Highlights in the history of Artificial Intelligence}
     \label{fig:AIHIST}
\end{figure}

We are currently witnessing an era of exponential success in the field of artificial intelligence that has been evolving for more than 50 years (See Figure \ref{fig:AIHIST}). 
Meanwhile, it is common knowledge that progress is slow in terms of innovation and scientific knowledge development in the field of AEC. 
 
\begin{figure}[htb!]
     \centering
     \includegraphics[width=\textwidth]{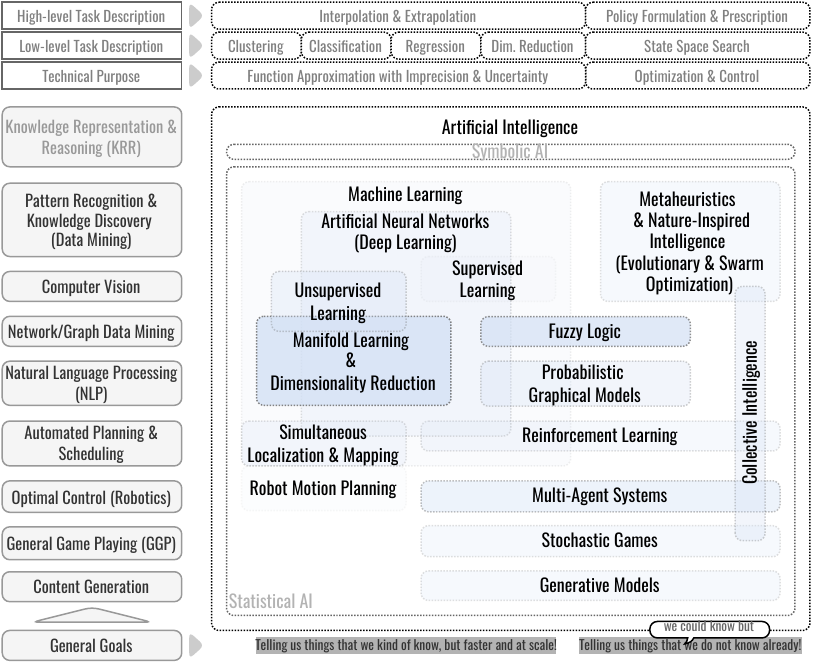}
     \caption{An Euler diagram of the scope of Artificial Intelligence}
     \label{fig:AI_Scope}
\end{figure}

As has been extensively argued before by \citein{simon_structure_1973} and \citein{azadi_godesign_2021}, once an unambiguous language is adopted for discussing the classification of problems, we can see that many of the problems of the AEC can be dealt with very adequately (and possibly painstakingly) through conventional mathematics, physics, and computer science.
In other words, the utility or the necessity of employing AI for dealing with problems that can be dealt with through conventional mathematical or computational procedures is not only pointless from a resource-efficiency stance but also questionable from the point of view of interpretability, transparency, and explainability. 
To assess the potential applications of AI in AEC w.r.t. these questions, we highlight the history of AI (see Figure \ref{fig:AIHIST}) and refocus on the scope of AI (see Figure \ref{fig:AI_Scope}), at least as could possibly pertain to AEC.

Once a problem is formulated adequately, two major determinants can be considered as to whether it would be sensible to apply AI or not: whether the data schemata of the problem are structured (vectorized) or unstructured (textual/visual), and whether the underlying associations between the inputs and outputs can be modelled through first principles (governing laws of physics, typically stated in differential equations), stochastic processes, or Agent-Based Models.
If the problem data are unstructured or the conventional modelling approaches do not have the capability of capturing the complex associations between the inputs and the outputs of interest, then, especially when interpretability can be sacrificed over the necessity of predictability, utilizing AI is quite sensible.
The example that we shall discuss in this chapter may seem somewhat questionable according to these points but on the other hand, it is too overwhelmingly large and complex that no conventional approach can deal with it at the aimed level of abstraction. 
In this case, the ambition of the project is on such a high level of abstraction for policy analysis that the inaccuracies and ambiguities of the purely data-driven approach can be justified because of the insights that can be gained from the meta-statistical model. 
\subsection{Design as Decision-Making}
The commonly overstated notions of difficulty or ill-defined nature of design problems, see e.g. \citein{simon_structure_1973}, can be attributed to the fact that most design tasks are expected to produce a very concrete geometric description of an object to be built (the form), given only a very abstract description of what the object is supposed to be used for, how it should work, and what would be desirable for it to achieve, all of which are often described quite vaguely (the function), q.v. \citein{kroes2006dual}. 


Hillier was one of the few shrewd theorists who understood that, at least after the separation of structural design from architecture in the 19th century, q.v. \citein{giedion2009space}, what distinguishes building buildings from architecture is the art and science of configuring spaces, as stated in ``Space is the Machine'', \autocite{hillier2007space}. 
Once one realizes that the so-called architectural form is not only a single shape of an iconic object but also a set that includes the shapes of spaces and eventually the constituent segments of a building then we can distinguish the superior importance of spatial configurations. 
As obvious as it may sound, it seems to be necessary to emphasize that architectural design is not merely about sculpting a shape but configuring spaces to accommodate some human activities.
This involves some puzzling tasks such as packing, zoning, and routing spaces of various functions, which we hereinafter refer to as the task of configuring buildings \autocite{azadi_godesign_2021}.
For the problems of shape and configuration to be transformed into decision problems, they need to be discretized rigorously. 
In short, we can call a massing problem a shape problem and a zoning problem a configuration problem (see Figure \ref{fig:ConfigSpace}).


The mainstay of the generative design paradigm is a rigorous reformulation of a design problem as a discrete topological decision problem rather than a geometrical problem \autocite{Nourian2023}.
Therefore discretization is the process of breaking down the integrated design problem into multiple smaller yet interdependent decision problems.
An example of such discretization can be a voxel grid that provides a non-biased and homogeneous representation of spatial units, each of which poses a decision problem of function allocation \autocite{nourian_voxelization_2016} \&\autocite{soman_aditya_decigenarch_2022}. 

Moreover, to ensure the correspondence of these discrete decisions we need to include the topological information about their neighbourhood to represent their spatial inter-dependencies; similar to topology optimization \autocite{oshaughnessy_topology_2021}. 
At the limit, such discretization can also be used to model a continuum of solutions and provide a frequency-based or spectral representation system, similar to \autocite{marin_spectral_2021}, for spatial design much like the musical notation that is based on notes. 

Additionally, it is important to note that design decisions have a strong spatial dimension, however, they can include the social dimension to represent the preference of stallholders and enable consensus-building \autocite{bai_decision-making_2020}.
Given a view of design as a matter of decision-making, we can readily see two important types of practical questions that will shed light on the relevance of AI for decision-making:
\begin{enumerate}
     \item how to map/learn the associations of some hundreds or thousands of constituent choices of a compound design decision (function approximation and dimensionality reduction for ex-ante assessment of the impact of decisions)?
     \item how to navigate a gigantic decision space with thousands of choices and their astronomically large combinations with a few important consequences in the picture?
\end{enumerate} 


The proposed notion of design as decision-making makes a point of departure for the rest of the chapter in that it highlights two essential problems of equal importance and significance that can be tackled by AI and their duality: 
Firstly, evaluation problems can be portrayed as mapping problems in Machine Learning and Deep Learning, where the approximation power of Artificial Neural Networks (ANN) can be exploited in regression and classification settings.
Secondly, derivation problems can be portrayed as navigation problems in generative models, concerned with navigating from a low-dimensional representation of performance indicators towards disaggregated design decisions.

\begin{figure}
     \centering
     \includegraphics[width=\textwidth]{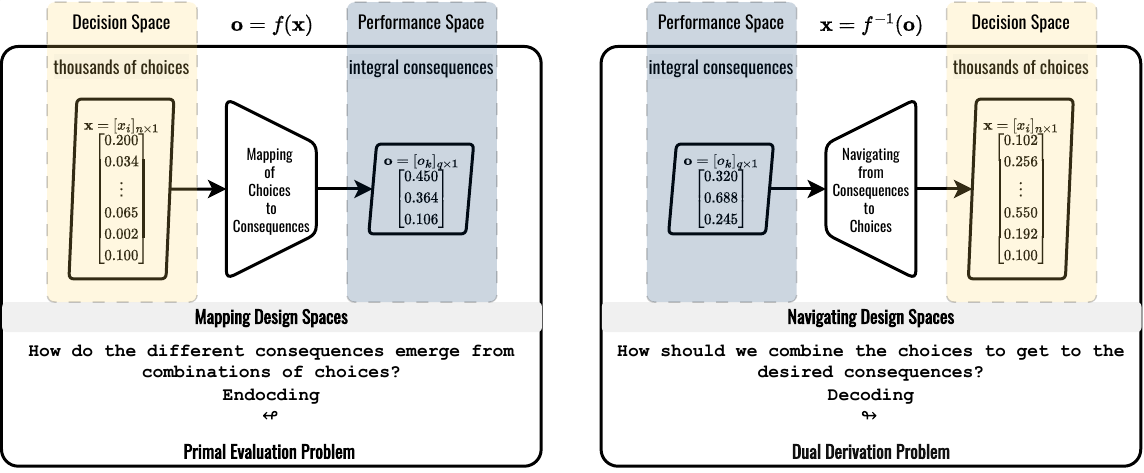}
     \caption{The duality of evaluation and derivation problems in generative design}
     \label{fig:AI_MapNav}
\end{figure}

Encapsulating the complex and often non-linear associations of many design decisions with a few outcomes of interest or performance indicators is here dubbed as a mapping problem. 
Inverting a map (as an approximated function, e.g. in the form of an ANN), can thus be viewed as an enhanced or augmented form of design, where the designer is navigated towards many small decisions just by pointing towards certain data points within a low-dimensional performance space (see Figure \ref{fig:AI_MapNav}). 
It must be apparent that a navigation problem in this sense is much harder to solve; almost always impossible in the absolute sense due to an arbitrarily large increase in information content and thus a combinatorial explosion of possibilities.

\subsection{AI for Generative Design}

\begin{figure}
     \centering
     \includegraphics[width=\textwidth]{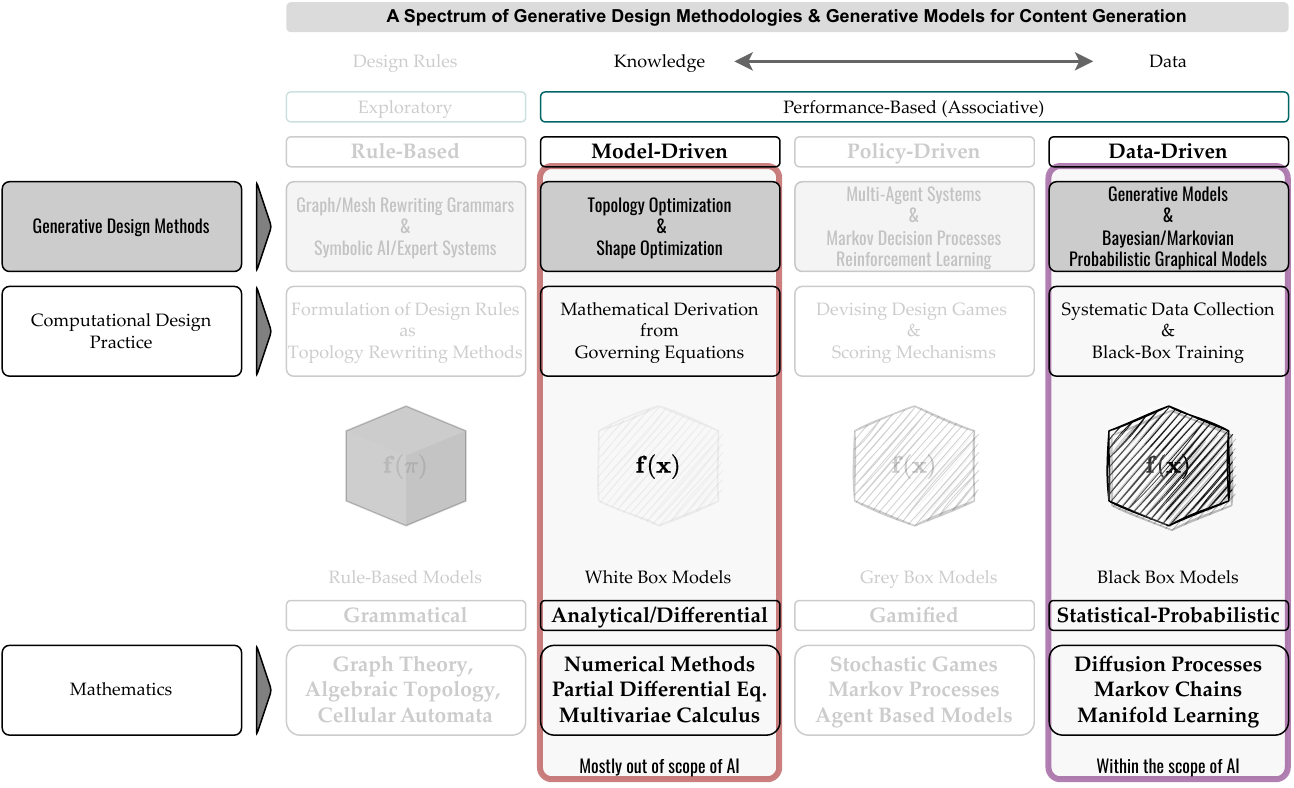}
     \caption{The spectrum of generative design methods and their relation to AI methods}
     \label{fig:AI_Generative}
\end{figure}

Given the formulation of main generative design tasks as \emph{mapping} and \emph{navigating}, we focus on a particular set of AI methods that are distinguished as to their relevance for these tasks in high-dimensional design decision spaces.
More specifically, within the spectrum of generative design methods \autocite{Nourian2023}, we focus on data-driven \emph{mapping} and \emph{navigating} strategies.
As shown in figure \ref{fig:AI_Generative}, for brevity, we will only focus on the data-driven approaches to design on the right-hand side of the spectrum.
Notwithstanding the other possible applications of [different kinds of] AI in this generative design spectrum, such as Reinforcement Learning in Policy-Driven design (playing design games), approximation of evaluation functions in topology or shape optimization, and Expert Systems in grammatical design, our framework here is focused on the statistical AI paradigm and so we only discuss the purely data-driven approaches to generative design. 

Two subtle issues must be noted here: 
firstly, instead of discussing the utility of the wondrous application of generative models for the entertainment industry, we shall reflect on how the generative processes based on diffusion or dimensionality reduction can be controlled for attaining high-performance designs in an explainable manner.
Secondly, the model-driven approaches to performance-based generative design (topology optimization in particular) based on first principles, are already utilizing something important from the realm of nature-inspired computing called Hebbian Learning, which is already in the scope of [statistical] AI. 
This point, however important, generally interesting, and relatively unknown, falls out of the scope of this chapter.  

%

\section{Framework} 
\label{sec:frame}
%


The emphasis on the decision-making approach to design entails that design tasks can be formulated as a set of [typically unstructured] questions about the form and materialization/construction of an object (a building) to be answered. 
In this chapter, we focus on the questions that pertain to form.

We propose a mathematical framework for generative design that relates multiple strands of work together.
We use \emph{design space} for referring to an ordered pair of two vector spaces: a decision space containing vectors or data points representing design configurations in the form of $\mathbf{x}\in (0,1]^n$ and a performance/quality space containing vectors or data-point representing combinations of outcomes of interest in the form of $\mathbf{o}\in [0,1]^q$.

\begin{figure}
     \centering
     \includegraphics[width=\textwidth]{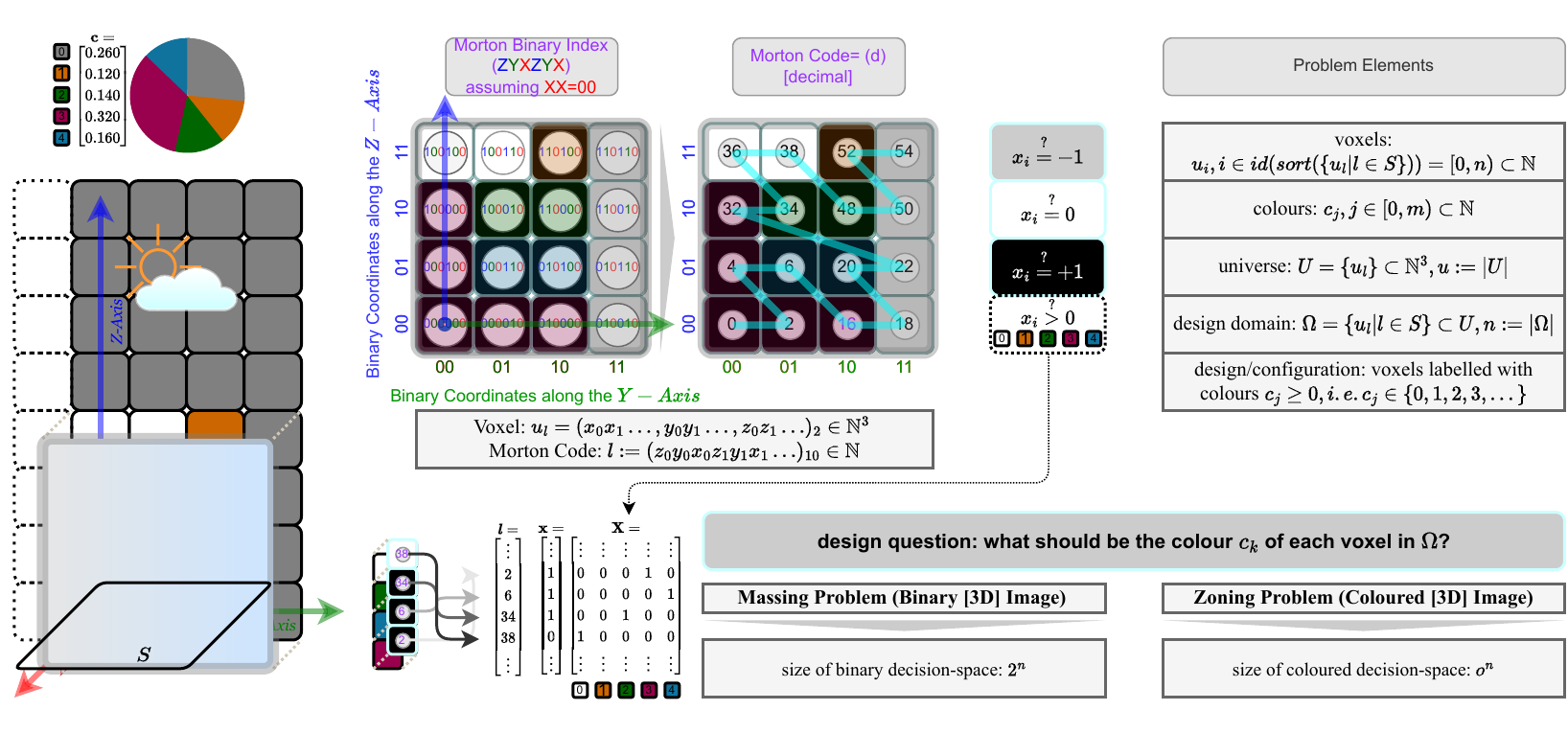}
     \caption{An illustrative discrete design domain and its associated decision space distinguished for shaping/massing and zoning/configuring problems}
     \label{fig:ConfigSpace} 
\end{figure}

The mathematical lens that we shall put on the issue is to redefine both of these notions to provide a much more specific and workable idea for discussing the utility or futility of applying AI to design problem-solving. 
It is hopefully easy for the reader to accept that a regular discretization of a so-called \emph{design space} (which is an unfortunately common misnomer, but here somewhat pragmatically useful) provides a straightforward and simple discretization of design decisions as vectors in the form of $\mathbf{x}:=[x_i]_{n\times 1}\in (0,1]^n$ or $\mathbf{x}\in \{0,1\}^n$, where $n$ is the number of discrete cells in the design space in which virtually any conceivable shape can be constructed at a certain level of resolution. 
Without loss of generality, however, the decision variables are not necessarily spatial and can be assumed to be relativized float variables in the range of minimum and maximum admissible parameter values of the functions that together result in the shape and configuration of a building. 
Even if a multi-colour (multi-label, multi-functional) space is the subject of the design problem, then multiple categories/colours of such vectors can be seen together as a matrix of decision variables, whose rows have to add up to 1 (see Figure \ref{fig:ConfigSpace}). 

Once this terminology is established, it is easy to observe that, in Machine Learning (ML) terms, the problem of performance-based design can be seen as two problems that are dual to one another, a multi-variate regression problem for figuring out an approximation function that can map a few outputs to many inputs (here we call this mapping or the evaluation problem), and a pseudo-inversion problem for finding the combination of inputs that could result in desired output data points (here we call this navigation or the derivation problem), see Figure \ref{fig:AI_MapNav}.

When approached as a data-driven problem-solving task, both problems are somewhat hard and impossible to solve in the absolute sense of the word, unless we think about them as loss minimization or approximation problems. 
The navigation/design problem is much harder than the mapping/evaluation problem. 
The main idea here is to advocate for training (fitting) meta-models (neural networks) to sets of sampled pairs of inputs and outputs to first approximate a complex design space as a map between decision data points and performance data points, and then find the pseudo-inverse of this map or navigate it in the reverse direction to be able to find designs (decision data points) that perform in a desired way. 
In other words, mathematically, we look at the performance-based design process as a pairing between a decision space and a performance space, where a map is conceptualized as a function $f: (0,1]^n\mapsto [0,1]^q)$ such that $\mathbf{o}=f(\mathbf{x})$.
The pseudo-inverse map is thus dubbed as $f^{-1}: [0,1]^q \mapsto (0,1]^n$, such that $\mathbf{x}=f^{-1}(\mathbf{o})$. 

For the sake of brevity and also generalizability to non-spatial design problems, we shall focus on massing problems and leave coloured configuration problems out of the picture momentarily (see Figure \ref{fig:ConfigSpace} for the distinction).  

Furthermore, by considering two abstract and high-level descriptions of a design task in our proposed regular discretization frameworks, we can formulate two mathematical tasks:
\begin{itemize}
     \item Mapping Design Spaces: approximating the function that can model the associations between the many input design variables and a few outcomes of interest;
     \item Navigating Design Spaces: approximating the inverse function that can guide the generation of valid configurations in the decision space given desired data points in the performance space.
\end{itemize}

\subsection{Design Space Exploration} 
Here we explain the mathematical meaning of the two dual problems that together can be called design space exploration tasks: mapping and navigation. 
\paragraph{Mapping} 

The problem of mapping associations between a large set of independent input decision variables and dependent output performance indicators is key to performance-driven design. 
Any explainable and accountable design methodology should have the capacity to guarantee the attainment of some quality or performance indicators. 
From a mathematical and statistical point of view, we might prefer to have an explainable and interpretable model of such relations that can be fitted into our data or ideally a simulation model to predict outputs from input data. 
However, in some cases, especially where a multitude of very different quality/performance indicators are involved, and when one does not have an established basis for simulation modelling, statistical (data-driven) modelling seems to be the only option.
And so, when the complexity of the model passes a certain threshold of non-linearity and a multitude of input outputs, we might prefer to trade interpretability for predictive power. 
That is exactly where ANNs as families of adjustable non-linear functions stand out as viable function approximators. 
Training a network is practically a matter of minimizing a loss/error function by adjusting the parameters of a family of functions (that is set out by the so-called architecture/structure of the ANN). 

Even though this approximation is inherently non-linear and global, it is illuminating to think of an alternative [locally] linear approximation based on the Jacobian Matrix. 
Suppose that $\mathbf{o}=f(\mathbf{x}):=[f_k(\mathbf{x})]_{q\times 1}=[f_k([x_i]_{n\times 1})]_{q\times 1}$ is a vector of multiple scalar functions of vector input variables. 
Then a basic idea of approximation is to approximate this function locally around an input data point by its Jacobian. 
This matrix operator gives the basis for a hyper-plane equation that provides the $n$-dimensional Euclidean tangent space of the underlying function, very much like a multi-variate regression hyperplane, albeit the latter would be fitted to the entire dataset. 

Note that the Machine Learning task here would be a multi-variate regression task in this case, i.e. predicting the dependent given the independent variables. 
To understand the difficulty of the mapping then consider that the Jacobian matrix $\mathbf{J}:=[J_{k,i}]_{q\times n}=[\frac{\partial f_k}{\partial x_i}]_{q\times n}=[\nabla^T f_q]_{q\times 1}$ would just provide the best local linear approximation of an otherwise globally non-linear map from $\mathbb{R}^n$ to $\mathbb{R}^q$, i.e. $n$ decision variables to $q$ quality criteria or performance indicators.

The Jacobian approximation is numerically computable provided the underlying function is smooth and differentiable. 
For brevity, as commonly done, we have omitted the fact that the Jacobian can be evaluated at a certain input data point and that it is expected to be the best linear approximation of the function in question in the vicinity of that point. 
If we abbreviate the notation for the Jacobian as such a functional, then we can denote the approximate linear function at any given data point as follows (as a first-order Taylor Series expansion): $\mathbf{o}(\mathbf{x})|_{x \sim x_o}\simeq\mathbf{J}(\mathbf{x_0})\left(\mathbf{x}-\mathbf{x_0}\right)$, or simply put, as $\mathbf{o}\simeq\mathbf{J}\mathbf{x}$, if we assume $\mathbf{x}$ to represent the vector of difference between the input data point with the centre of the neighbourhood. 

The Jacobian approximation is also illuminating for another important reason: it allows us to approximate the Jacobian in a different sense, i.e. in the sense of dimensionality/rank reduction using the Singular Value Decomposition (SVD) to see a clearer picture of the main factors playing the most significant roles in attaining the outcomes of interest, in other words identifying the inputs variables to which the outcomes of interest are most sensitive. 
Even though we do not explicitly perform this operation in our demonstrative example using the SVD, it is still illuminating to see what SVD can do for this insightful approximation and dimensionality reduction for two reasons:
\begin{itemize}
     \item The SVD approximation of the Jacobian allows us to make a cognitive and interpretable map of the most important causes of the effects of interest
     \item The SVD approximation of the Jacobian allows us to conceptualize a pseudo-inverse function to navigate the design space from the side of performance data points. 
\end{itemize} 

The SVD (low-rank) approximation of the Jacobian Matrix can be denoted as below:
\begin{equation}
     \mathbf{J}:=\mathbf{U}\boldsymbol{\Sigma}\mathbf{V}^T,
\end{equation}
where $\mathbf{U}_{q\times q}:=[\mathbf{u}_k]_{1\times q}$ and $\mathbf{V}_{n\times n}:=[\mathbf{v}_i]_{1\times n}$ are orthogonal matrices (i.e. $\mathbf{U}\mathbf{U}^T=\mathbf{I}_{q\times q}$ and $\mathbf{V}\mathbf{V}^T=\mathbf{I}_{n\times n}$), and $\boldsymbol{\Sigma}$ is a matrix of size $q\times n$ with only $p=min\{q,n\}$ non-zero diagonal entries denoted as $\sigma_c$ and called singular values, which are the square roots of the eigenvalues of both  $\mathbf{J}^T\mathbf{J}$ and $\mathbf{J}\mathbf{J}^T$, sorted in descending order, q.v. \autocite{martin2012extraordinary}. 
\begin{equation}
     \mathbf{J}\simeq\sum_{c\in[0,r)} \sigma_c \mathbf{u}_c\mathbf{v}_c^T, 
\end{equation}
where, $r\leq p$. It must be noted that the sum is not meant to be exhaustive in that the sum of the first significant terms would achieve the purpose of dimensionality reduction of the decision space by showing a low-dimensional picture of the correlations between decision variables and their performance consequences. So, instead of decomposing the Jacobian up to $p$, we can choose to have a lower dimensional approximation up to some arbitrary smaller number $r$. 


\paragraph{Navigation}

Navigating a high-dimensional design space from the side of the performance space towards the decision space for deriving design decisions (see Figure \ref{fig:AI_MapNav}) is clearly a very challenging task, almost always impossible in the absolute sense of solving the equation $\mathbf{J}\mathbf{x}=\mathbf{o}$, if the decision variables $\mathbf{x}$ are the unknowns. 

It is easy to see that the Moore-Penrose pseudo-inverse of the approximated Jacobian matrix can be computed as a matrix of size $n\times q$ by easily using the SVD factorized matrix:
\begin{equation}
     \mathbf{J}^{\dagger}:=\mathbf{V}\boldsymbol{\Sigma}^{\dagger}\mathbf{U}^T,
\end{equation}
where $\boldsymbol{\Sigma}^{\dagger}$ is simply formed as a diagonal matrix of size $n\times q$ with the reciprocals of the singular values. 
Similarly, the approximate pseudo-inverse of the Jacobian can be computed as:
\begin{equation}
     \mathbf{J}^{\dagger}\simeq\sum_{c\in[0,r)} \sigma_c^{-1} \mathbf{v}_c\mathbf{u}_c^T.
\end{equation}
However, in the same way, a minimal loss approximate solution exists for such equations when the matrix is rectilinear, $\mathbf{J}^{\dagger}\mathbf{o}$ is expected to be the least-square solution to the linearized Jacobian approximate of a navigation problem. 
Even though the system might in theory have a solution, the odds of finding a unique solution are practically very skewed towards having an indeterminate system with many more inputs than outputs, and thus the system will have many approximate solutions rather than a unique exact solution. 

This is of course in line with the intuition of most human beings about the inherent difficulty of design problems for which there is no unique solution. 
Note that in all these theoretical treatments we implicitly assumed that all data points within the decision space correspond to valid designs whereas in reality, it might be more difficult to ensure finding valid solutions (feasible in the sense of complying with constraints) rather than good solutions. 
In other words, constraint solving tends to be more difficult than optimization in a feasible region of decision space. 

\subsection{Spatial Design Variables}
If the question of the design problem directly pertains to the shape of the configuration of an object, we can still construct decision variables to be handled within the proposed framework for mapping and navigating design spaces. 

The idea of bringing spatial decision variables in a generative design process is to consider first the nature of the objects being designed as manifolds, i.e. locally similar spaces (homeomorphic) to Euclidean spaces of low dimensions (2D planes or 3D hyper-planes) but globally more complex, possibly having holes, handles, and cavities (shells). 
Three types of these manifolds are of special interest for generative design, such as those that conduct walk flows (explicit or implicit pedestrian corridors in buildings and cities), light flows (rays of sunlight, sky-view, or other visibility targets), and force flows in structures.
Our conceptual framework proposes that these flows are conducted within spatial manifolds as below:

\begin{enumerate}
     \item Walkable Space Manifolds (2D) conduct walks (accessibility questions)
     \item Air Space Manifolds (3D) conduct light rays (visibility questions)
     \item Material Space Manifolds (3D) conduct forces (stability questions)
\end{enumerate}


This consideration allows us to see the way this object is supposed to function would largely be determined by how this manifold is configured in that the way the manifold in question conducts the flows of walks in a walkable floor space, flows of light rays in a visible air space, or flows of forces in a reliable material space. 
Thus we can highlight the specific concept of flow in a network representation dual to the discrete representation of a manifold as an unambiguous alternative intermediary instead of any vague notion of function to study and measure. 

Apart from mathematical elegance, this approach also provides multiple computational advantages that are very much in line with the recent advancements in the field of generative models in AI.
In a nutshell, the discrete representation of the so-called design space not only provides a workable representation of all possible forms but also a workable representation of some inherent functional properties of the represented manifold that should logically determine how it could function as a building (or a structure). 
The manifold representation can be mathematically denoted as a polygon mesh of vertices, edges, and faces $\mathcal{M}=(V,E,F)$ (for a 2-manifold) or a polyhedral mesh of vertices, edges, faces, and cells $\mathcal{M}=(V,E,F,C)$ that can have a dual graph representation in the form of $\Gamma=(N,\Lambda)$. 

This description should principally sound natural if we articulate the purpose of a design task as follows: finding the ideal form (configuration and shape) of a manifold to conduct some flows in a desirable pattern. 
In this way, we are diverting our attention from the containers of space (the building) into what it contains (i.e. the space and its spatial configuration). 
This change of focus allows us to see the direct correspondence between the so-called form and function of a design or, better put, the form (i.e. configuration \& shape) and expected quality/performance of a spatial configuration. 

In what follows, we shall go much beyond the vector data inputs consisting of only numerical variables, especially in the context of our illustrative example. 
In fact, without loss of generality, the ideas of mapping and navigating design spaces in an approximate sense go beyond decision variables pertaining to continuous decision variables and those pertaining to the spatial configuration and geometric shape of spatial manifolds. 
The same ideas can be applied to design problems that are about decision-making in a much more general sense as discussed above. 
Note that the illustrative example that we have demonstrated at the end of the chapter has a heterogeneous mix of spatial and mostly non-spatial decision variables as well as a mix of categorical and numerical decision variables.

\subsection{Statistical Approaches to Design}


Amongst the statistical approaches to design, we can distinguish the possibilistic approaches from the probabilistic ones. 
\begin{itemize}
     \item probabilistic approaches: BBN, Variational Auto-Encoders (VAE), and Diffusion Models
     \item possibilistic approaches: Markovian Design Machines, Fuzzy Design (see MAGMA below)
\end{itemize}

\paragraph{Possibilistic Approach}
The essence of the possibilistic approach to design is using a multi-valued or non-binary logic framework for making design decisions, typically in the sense of making discrete choices about discrete segments of space, for example, the Markovian Design Machines of \autocite{batty_theory_1974}, the Spatial Agents Academy of \autocite{veloso_academy_2020}, and MAGMA (Multi-Attribute Gradient-Driven Mass Aggregation) through Fuzzy Logic as introduced briefly in \citein{nourian_configraphics_2016} and \citein{soman_aditya_decigenarch_2022}. 
Both of these methodologies apply non-binary logic from a possibilistic point of view, in the sense that they take design inputs that are valued in the range of $[0,1]$ but treat them as possibility measures rather than probability measures. 
The two big ideas behind these two methods are the utilization of Markov Chains, Markov Decision-Processes, and Fuzzy T-Norms for coping with uncertainty and human-like reasoning in simulated negotiations between spatial agents. 

\paragraph{Probabilistic Approach}
The probabilistic models briefly mentioned here are all related to the concept of conditional probability, the Bayes theorem, and [generalized] stochastic processes that resemble Markov Chains \autocite{weng2021diffusion,nourian_configraphics_2016}. 
In a nutshell, the core of these models is about updating some posterior probabilities indicating beliefs about the truth of some statements by prior probabilities multiplied by the likelihood of compelling evidence, scaled by the probability of the existence of the evidence. 
When probabilistic neuron-like nodes in Probabilistic Graphical Models are combined, these new posterior probabilities or probability distributions can be fed into other layers of a network to create ANN architectures. 
A basic idea here is to gradually reduce the dimensionality of input data into an abstract low-dimensional representation (encoding, or mapping, albeit into a typically unintuitive and interpretable latent space) and then gradually use the inverse of the forward diffusion-like processes to denoise a vector in the low dimensional hidden space. 
The latter process is called denoising or decoding and it matches our description of navigation processes, albeit without the direct control of the meaning of the latent space vectors. 
A breakthrough in this domain can come from enhancing the explainability of the latent space low-dimensional representations. 
This idea, however interesting, falls way outside the scope of this short treatise. 
Therefore, here we only provide a theoretical minimum for understanding the demonstrative example (i.e. a shallow Bayesian Belief Network).

\section{Demonstration}
\label{sec:demo}

In this section, we will present a demonstration of the utility of the proposed framework to indicate how a discrete decision-making approach can facilitate generative design processes.
As a disclaimer, it must be noted that this example is not chosen for technical reasons related to AI but rather due to its real and societal and environmental important purpose for policy analysis concerning energy transition planning actions at the level of a country, and sustainability strategies at the building level. 

\subsection{Case Study}

Understanding the energy performance of architectural designs is crucial in ensuring a sustainable future.
Building Energy Modelling (BEM) is a multi-purpose approach used by designers and policymakers for checking building code compliance, certifying energy performance, subsidy policy making, and building management.
The Dutch government has recently introduced the NTA 8800 calculation model for quantitatively determining the energy performance and code compliance of buildings \autocite{gebouwenergieprestatie2022}. 
The NTA 8800 aims to provide a transparent, verifiable, and enforceable building energy performance model, based on the European Energy Performance of Buildings Directive (EPBD), the European Committee of Standardization (CEN), and the Dutch Normalization Institute (NEN) published standards \autocite{nen_2017}.
These regulations describe methods to calculate the energy performance of buildings, set energy requirements for new buildings, and make agreements about energy label obligations in existing buildings.
The NTA 8800 only concerns building-related measures, as expressed in the EPBD, Annex A \autocite{europian_union_2021}. 

The NTA 8800 document has been implemented as an MS Excel tool by the Dutch government (commissioned by Nieman B.V. consultants). 
This calculation model translates the public European standard document into a calculation tool.
The calculation tool is not publicly available and it is not documented. 
Since we were given temporary and bounded access to this model we chose to approximate it and construct a meta-model out of it. 
The model consists of 269 unique input parameters about the spatial and technical building design configurations, based on which the model returns three scalar response values about the energy performance of the building design; BENG 1 (maximum permissible energy demand in $kWh/m^2y$), BENG 2 (maximum permissible primary energy consumption in $kWh/m^2y$), and BENG 3 (minimum permissible share of renewable energy use as a percentage).
The acronym BENG refers to some national performance indicators for Nearly Zero-Energy Buildings (Bijna Energie Neutrale Gebouwen in Dutch).

NTA 8800 model has three main limitations: (1) it can only process and compute information about a single specific scenario at a time; (2) it returns scalar values about the energy performance that is untraceable to input parameters; and (3) missing input values could result in errors or non-realistic response values.
These three limitations make the model impractical for designers and policy analysts, particularly in the early stages of design.
This impracticality is because, in conceptual design and policy analysis, designers need to (1) explore and iterate various options simultaneously; (2) need feedback on the degree of influence of each design decision; and (3) cannot provide detailed information yet about later stage design choices, such as the technical systems.

The framework of Augmented Computational Design (ACD) is particularly useful here as it allows us to relate the aggregated performance changes of the few NTA 8800 outputs of interest to the changes in the many design decision parameters of its input.
In this particular case, we adopt a probabilistic meta-modelling (function approximation) approach based on the methodology suggested by \autocite{conti2021explainable}.

\subsection{Methodology}


\paragraph{Meta-Modelling}
Meta models are models that describe the structure, behaviours, or other characteristics of related models, providing a higher-level abstraction for constructing and interpreting complex numerical models that approximate more sophisticated models often based on simulations. 
A meta-model is to serve as a simplified, computationally efficient \emph{model of the model} \autocite{conti2021explainable}, also referred to as a surrogate model \autocite{Kleijnen1975ACommentBlanning}.
The process of creating a meta-model is referred to as meta-modelling \autocite{van1991system}.
Some alternative meta-modelling techniques include interpolation methods such as spline models \autocite{BartonSimulationMetamodels}, polynomial regression \autocite{KleijnenLowOrderPolynomial}, or Krigging \autocite{Ankenman2010StochasticKrigingFor}. 

Within the ACD framework, such meta-models provide structured ways to perform the two most important tasks of the generative design: \emph{mapping} and \emph{navigation}.


In general, a standard meta-model can be described as: $\mathbf{o}=\mathbf{f}(\mathbf{x})\simeq \mathbf{g}(\mathbf{x})$, where $\mathbf{o}$ is the aggregated simulation response, $\mathbf{f}$ denotes a computational simulation-based model conceptualized as a vector function and $\mathbf{g}$ is the approximated model function (cf. Figure \ref{fig:AI_MapNav}.)
With this notation, the objective of meta-modelling is to build the $\mathbf{g}$ in such a way that it produces reasonably close values of $\mathbf{o}$.
In the case of augmented computational design, meta-modelling can be adopted as a methodology of design \emph{mapping} that provides a differentiable and ideally reversible $\mathbf{g}$ that can be used in the \emph{navigating} process.
In other words, the meta-modelling should structurally relate the choices and consequences in such a way that the choices can be derived from the desired consequences; hence providing a data-driven basis for generative design. 
The next part demonstrates a probabilistic meta-modelling approach to navigation tasks in high-dimensional design decision spaces, based on the methodology introduced by \citein{conti2021explainable}.

\paragraph{Bayesian Belief Networks}
A Bayesian Belief Network (BBN) is a kind of Probabilistic Graphical Model (PGM) that is effectively an ANN in the form of a Directed Acyclic Graph (DAG) with neuron-like nodes that can compute Joint Probability Distributions (JPDs) from input probability distributions or discrete Probability Density Functions (PDF), which is then attributed to an output probability distribution through a Conditional Probability Distribution (CPD) computing posterior probabilities/beliefs through the Bayes theorem, hence the name Bayesian. 
The set of edges of a BBN forms the model architecture or structure that represents the particular probabilistic dependencies between the discrete probability distributions attributed to the starts and ends of the nodes.
This structure is typically set by the modeller based on their knowledge of the process; while the conditional probability distributions (transition probability matrices) are learnt from the experimental data.  
BBNs can help us semi-automatically reason about uncertain knowledge or data \autocite{PENG2010BayesianNetworkReasoning}.
This makes it possible to perform probabilistic inference, such as computing the JPDs of some outputs (effects) given some inputs (causes).
The name of these ANN comes from the idea of updating beliefs or hypotheses posterior to observing evidence; more precisely, utilizing the Bayes theorem for updating conditional probabilities in network structures, in a fashion similar to modelling and evaluating Markov Chains, albeit with the difference that Markov Chains operate as uni-partite networks but each neuron of a BBN is a bipartite network coupled with an outer product calculator. 

The neurons of a BBN consist of two layers, the first of which can be dubbed a presynaptic layer that combines input discrete probability distributions (through an outer product) and forms a JPD and then flattens the JPD to form a vector-shaped probability distribution. 
The second, i.e. the synaptic layer then is a CPD, i.e. practically a rectangular probability transition matrix that maps this flattened JPD to the output probability distribution. 
A BBN then consists of such neurons connected in a DAG. 
Training a BBN means finding the entries of the CPD in such a way as to minimize the loss of recovery of the output probability distribution from the input distributions. 
The appeal of BBNs is twofold: on the one hand, they allow the inclusion of expert knowledge and intuition in the architecture of the network and on the other hand the training of the network makes the network adapted to the objective data.
In this case, we limit the architecture of the network to a single layer of neurons to keep the network invertible. 

\begin{figure}[ht]
     \centering
     \includegraphics[width=\textwidth]{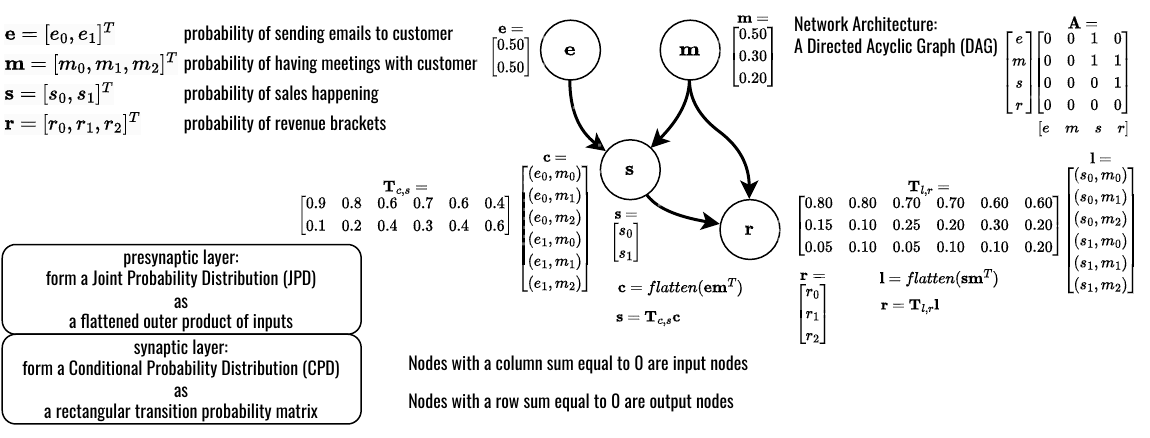}
     \caption{An illustrative example of a Bayesian Belief Network, eliciting the nature of nodes and the network architecture, an example adapted from \href{https://causalnex.readthedocs.io/en/latest/04_user_guide/04_user_guide.html}{here}}
     \label{fig:AI_BBN}
\end{figure}

\begin{figure}[ht]
     \centering
     \includegraphics[width=\textwidth]{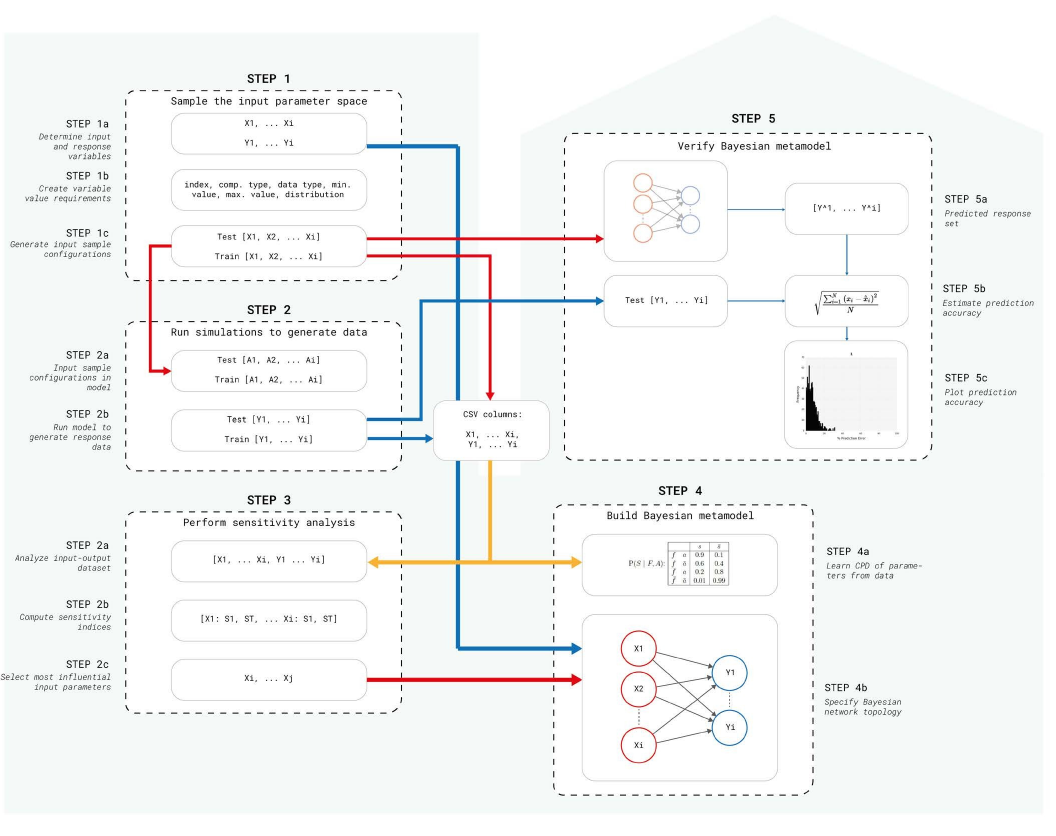}
     \caption{Overview of the workflow adopted from the framework introduced by \autocite{conti2018FlexibleSimulation}}
     \label{fig:method}
\end{figure}

\paragraph{Workflow}
Research from \citein{conti2018FlexibleSimulation} illustrates the four process steps involved in developing a BBN meta-model.
In this use case, we alter this methodology as creating a BBN with all 269 input parameters is infeasible.
We add an intermediary step of sensitivity analysis to identify the most influential input parameters before constructing the BBN.
Thus we follow these steps in order (see Figure \ref{fig:method}): (1) sample the input parameter space, (2) run simulations to generate the output values, (3) sensitivity analysis and selection of influential input parameters, (4) train the BBN, and finally (5) evaluate the model's robustness. 
As \citein{conti2021explainable} highlights, it is important to model a shallow BBN as a complete bipartite graph connecting all input nodes to all output nodes, effectively limiting the topology to two layers.
This would allow us to make a reversible approximation that can be used to derive the necessary input configuration for any desired performance output.
Additionally, the fixed values can also include some of the input variables turning them into design constraints.

\paragraph{Step 1: Sampling the Parameter Space}
We need to set up a Design of Experiment (DoE), to generate simulation data to study the relationships between various input variables and output variables \autocite{Hicks1964_HICFCI}.
The experiment involves running several simulations at randomised input configurations \autocite{Sacks1989DesignAnalysisComputer}. 
Before running the simulation, it is important to carefully select a sampling method, to determine these input configurations, since the chosen strategy influences the quality of the meta-model \autocite{Fang2005DesignModelingFor}.
Since it is assumed that the decision space is unknown, the intention is to be inclusive of all regions of the decision space as possible. 
The sampling algorithm should generate a well-varied response data set that captures all the information about the relationships between the input parameters and the responses. 
For this study, 20.000 quasi-random input samples were generated based on Sobol's sequences \autocite{sobol1990sensitivity} to ensure the homogeneity of the samples. 

\paragraph{Step 2: Run NTA 8800 Simulation Model}
Vectorization is an important part of the ACD, we represent the decision variables and outputs of interests as vectors (See Section \ref{sec:frame}.)
Each sample point can be interpreted as a vector of scalar input values $\mathbf{x}$.
Each batch of such vectors is fed to the NTA 8800 model to generate the vector of corresponding building performance outputs $\mathbf{o}$.
After running the primary simulation model for the sampled input data points, the response data is collated and linked to the input samples to form an input-output dataset for regression modelling (as in Machine Learning).

\paragraph{Step 3: Sensitivity Analysis} 
The creation of a meta-model from 269 parameters, each with scalar input values, requires a simulation of all possible combinations (the number of options to the power of 269). 
Even limiting the number of options for each parameter to two, results in an immense number of possible combinations, calculated at $5.39 \times 10^{80}$.
To contextualize the magnitude of this number, it is more than the estimated number of atoms in the observable universe.

The sheer magnitude of this number makes storage and training of Bayesian Belief Networks (BBNs) infeasible. 
Therefore, in this study, we use global sensitivity analysis to apportion the uncertainty in outputs to the uncertainty in each input factor over their entire range. 
This allows us to remove the parameters with the lowest influence on energy performance. 
The sensitivity analysis method is implemented in the workflow based on the SALib library \autocite{Herman2017SalibOpenSource}.

This results in a meta-model with fifteen parameters instead of 269, making it feasible to store and train the BBN, but on the other hand, reducing the accuracy and scope of the model. 
However, the most influential fifteen parameters are responsible for 90,45\% to 92,30\% of the final energy performance score. 
Hence, we decided on the inclusion of the fifteen specific parameters to construct the BBN meta-model.

\paragraph{Step 4: Build a BBN meta-model} 
Building a BBN meta-model is a process of associating the probabilistic relationships of inputs and outputs. 
These relationships may be characterized by a high degree of non-linearity and possibly multiple interactions and correlations between model parameters. 
Consequently, there are two main steps in this process: (1) learning the network topology as a DAG structure and (2) estimating the CPD attributed to the neuron-like nodes of the network.

\begin{figure}[h]
     \centering
     \includegraphics[width=.8\textwidth]{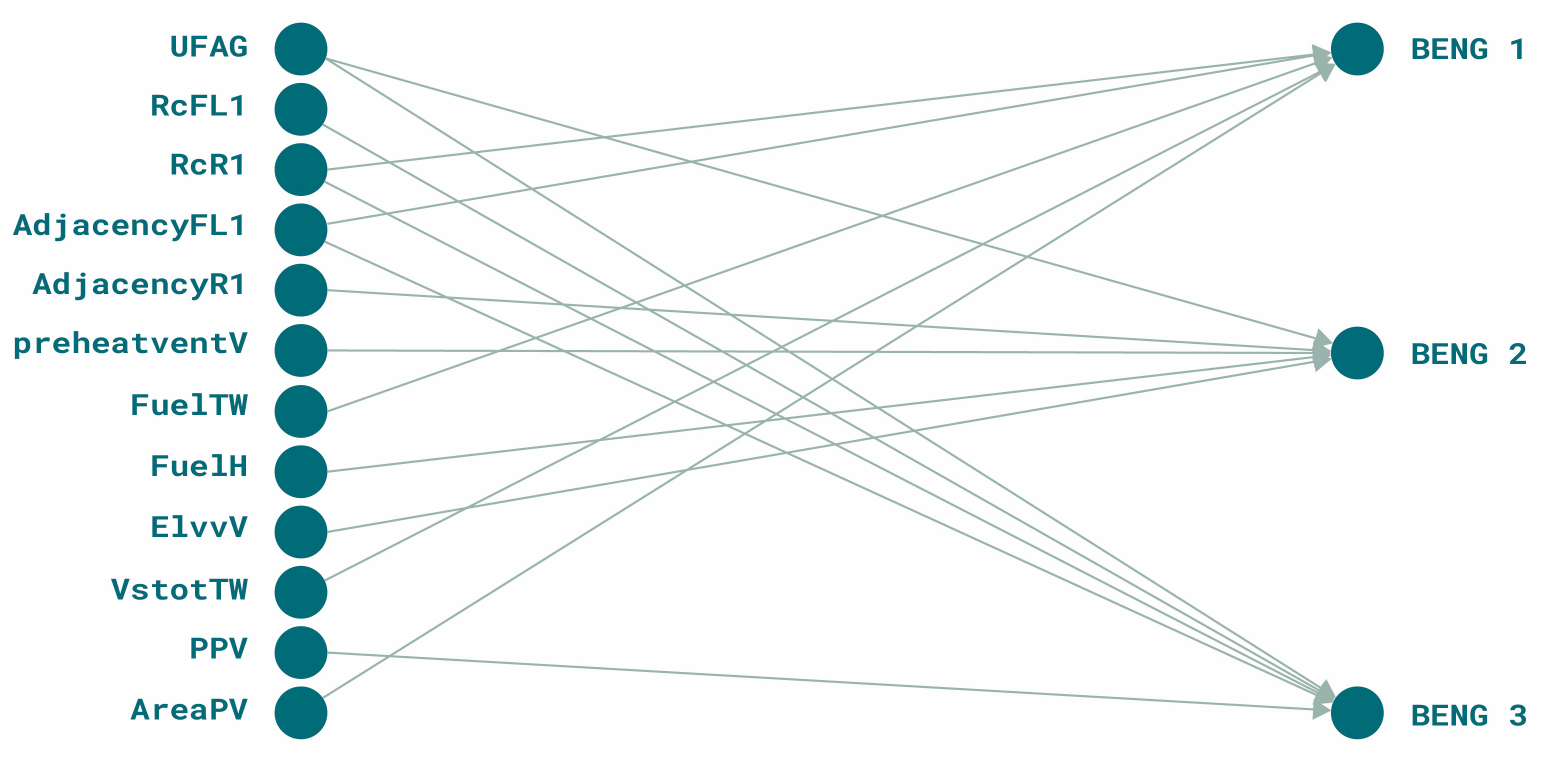}
     \caption{The single layer BBN; Right: BENG 1, BENG 2, and BENG 3 parameters are the output nodes; Left: Most sensitive input parameters as input nodes.}
     \label{fig:bbnsingle}
\end{figure}

In this demonstration, we adhere to a particular network topology to ensure the reversibility of the trained model \autocite{conti2021explainable}.
Accordingly, this BBN has only two layers: one corresponding to the input and one corresponding to the output.
However, effectively only a single layer of neurons operates in the middle of the two layers.
In this case, the selected parameters from the sensitivity analysis results are represented by the input nodes, and the BENG 1, BENG 2, and BENG 3 parameters are the output nodes (see Figure \ref{fig:bbnsingle}).
Therefore, we skip the topology learning step in the conventional BBN modelling; because the topology of this particular network is assumed to be a complete bipartite DAG.
In particular, we use the \emph{pgmpy} Python package to model the network topology \autocite{ankan2015pgmpy}.

The next step is to estimate the nodes' CPDs from the input-output dataset. 
The CPDs for the nodes can be learnt directly from the input-output simulation data generated in steps 1 and 2, using the Maximum Likelihood algorithm. 
Additionally, we discretize each variable range into a fixed number of intervals. 
All numerical input distributions generated using a space-filling approach, like Sobol's sequence or Latin Hypercube, are sampled based on continuous ranges, and should therefore be discretized. 
Discretization is done by dividing the interval of the parameter over a fixed number of ranges between the minimum and maximum values.

\paragraph{Step 5: Validating the Meta-Model} 
To assess how our trained BBN approximates the original NTA 8800 model, we use a cross-validation approach in combination with Normalized Root Mean Square Error (NRMSE) and Mean Absolute Percentage Error (MAPE) \autocite{james2013introduction}.
Cross-validation splits the generated input-output dataset (step 2) into a \emph{training set} and \emph{testing set}, before building the BBN (step 3). 
The BBN is trained on the \emph{training set} and assessed based on the \emph{testing set}.
However, to obtain a more reliable estimate of the model's performance, the dataset is split into several subsets or folds, with each fold used as both a training set and a testing set.
This research adopts a k-fold cross-validation technique, where $k$ refers to the number of groups that the data set is split into. 
We set $k=10$ based on experimentation to ensure a low bias and a modest variance.
The model is then trained on $k-1$ of the folds, and the remaining fold is used for testing. 
This process is repeated $k$ times, with each fold used for testing once. 
The performance of the model is then evaluated by averaging the performance across all $k$ runs. 

Following the approach suggested by \citein{conti2021explainable}, we computed the mean difference of the predicted and actual output values and normalize the RMSE values through division by standard deviation to achieve NRMSE.

To calculate the accuracy of the meta-model, it is recommended to use multiple metrics to get a comprehensive evaluation of the model's accuracy. 
Hence, NRMSE is combined with the MAPE metric. MAPE measures the average absolute percentage difference between the predicted and actual values. 
It is a measure of the magnitude of the errors in the model's predictions. 
Lower NRMSE and MAPE values indicate better model performance.
The larger the error between the two, the higher the NRMSE and MAPE values will become. 
Therefore, the NRMSE and MAPE results will indicate how dispersed the prediction data is compared to the actual model response.

\subsection{Results}
This section presents the numerical results obtained from the experiment of NTA 8800 meta-model. 
\subsubsection{BBN Validation Results}

Here we elaborate on the results of the cross-validation technique in combination with NMSRE and MAPE based on the test data set (s=1100).
The interpretation of what is considered an acceptable NRMSE and MAPE score depends on the specific problem and the context in which the meta-model is being used.
In general, it is recommended to compare the NRMSE and MAPE scores of the meta-model with the baseline models and the state-of-the-art models in the field. 
This can provide a benchmark for what is considered acceptable performance in the specific context of the problem.

In our case, the BBN does not compete with other models but rather competes with consulting building energy specialists in estimating the building energy performance in the early design stages. 
However, to assess the proficiency of our model in capturing the underlying relationships using solely the 15 selected parameters, we employ the following benchmarks: 
The NRMSE values should be in the range of $(0.20 \%, 0.60 \%)$ as baseline, and in the range of $(0.10 \%, 0.30 \%)$ as state-of-the-art \autocite{Bui2021MultiBehaviorWith};  
The MAPE values should be in the range of $(0.10, 0.30)$ as baseline, and in the range of $(0.05, 0.15)$ as state-of-the-art \autocite{Khan2021EnsemblePredictionApproach}. 

\begin{figure}
     \centering
     \includegraphics[width=\textwidth]{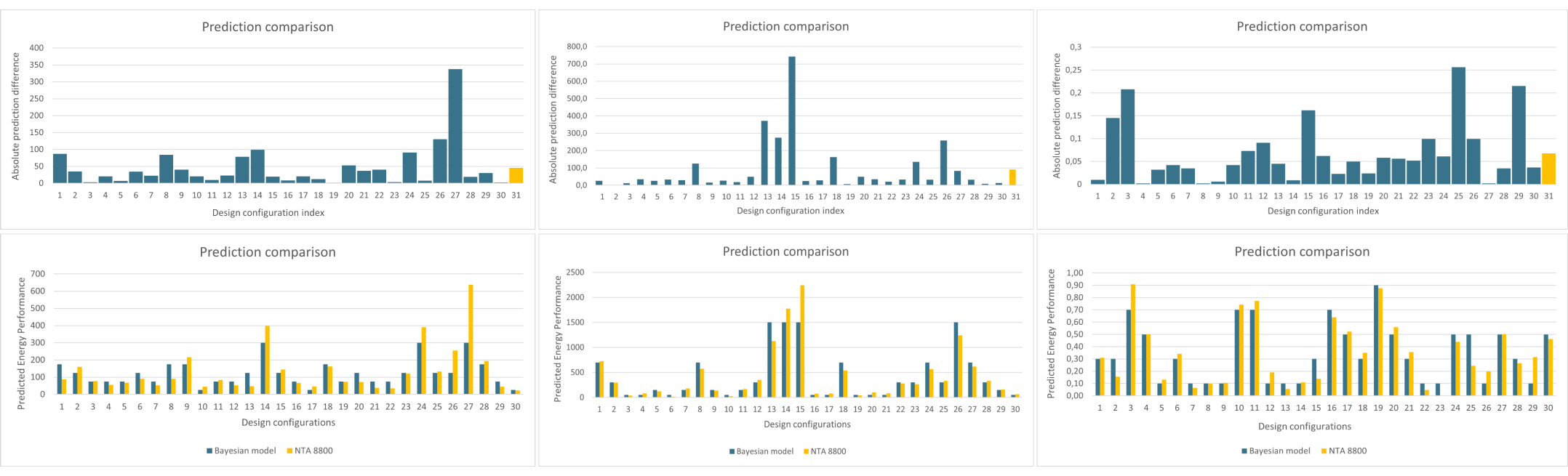}
     \caption{Histogram of BBN prediction and NTA 8800 outputs (top) and their comparison (bottom) for BENG1, BENG2, and BENG3, in order from lft to right.}
     \label{fig:prediction_comparison}
\end{figure}

The NRMSE of the trained BBN for BENG 1, BENG 2, and BENG 3 respectively is $0.82 \%$, $1.52 \%$, and $0.47 \%$.
This indicates that except for the BENG 3 indicator, the prediction of the model is not accurate enough.
The MAPE of the trained BBN for BENG 1, BENG 2, and BENG 3 respectively is $0.35$, $0.28$, and $0.33$.
This indicates that the predictions of the model are on the upper threshold of being acceptable as baseline models.
Absolute prediction difference can be seen in Figure \ref{fig:prediction_comparison}.

The prediction accuracy is calculated by $PA = \frac{100 y'}{ 100 * y - y'}$ Where $PA$ is the prediction accuracy, $y'$ is the prediction of the meta-model and $y$ is the NTA 8800 estimation.


\subsubsection{Toy Problem} 

Here we present a test case that demonstrates the effectiveness of the BBN meta-model in building design.
The study involves two toy problems that showcase the advantages and utilization of the meta-model. 
The toy problems regard two common design challenges that cannot be solved using the currently available tools, such as the NTA 8800. 
The first problem involves predicting the BENG 1 energy performance of a typical Dutch dwelling during the early design stage. 
The spatial characteristics of the building are fed to the meta-model.
As output, the meta-model returns a range and the confidence level of that range.

\begin{figure}[ht]
     \centering
     \includegraphics[width=0.5\textwidth]{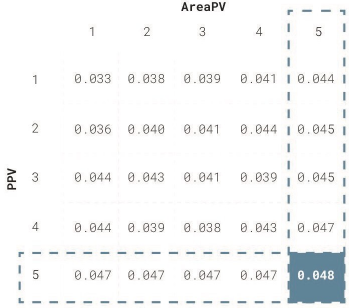}
     \caption{Output Recommendations of the BBN Meta-model for Achieving BENG 3 Compliance.}
     \label{fig:toy}
\end{figure}

In this toy problem, the meta-model predicts the BENG 1 value to be within the range of $(0-50) kWh/m^2.y$, with a confidence level of $100\%$ (see Figure \ref{fig:toy}).
To validate this result, we cross-checked the predicted result with the final configuration of the dwelling using the original NTA 8800 model. The NTA 8800 model returns a value of $39.8 kWh/m^2.y$, confirming the prediction capability of the meta-model.

The second problem reverses the first problem and involves the ex-ante determination of the most probable design configuration that satisfies a specific energy performance goal. 
In this example, the BENG 3 value of a typical Dutch dwelling design $(35\%)$ does not satisfy the minimal requirements $(50\%)$. 
Since this problem arises in the final design stage, some input parameters cannot be changed anymore.
In this case, the architects and engineers are limited to modifying only the area (AreaPV) and power (PPV) of the PV panels. 
Since the minimum required performance goal for BENG 3 is $50\%$, we set the goal value to a range of $60-80\%$.
Given the binning approach employed, it should be noted that the AreaPV value of 5 depicted in the figure corresponds to a range of $[40, 50] m^2$, while the PPV value of 5 corresponds to a PV Power range of $[200, 250] W/m^2$. 
Accordingly, the meta-model advises increasing the PV area to $[40, 50] m^2$, and the PV power to $[200, 250] W/m^2$ (See Figure \ref{fig:toy}). 

This discretisation allows for a clearer representation of the recommended parameter values within the specified ranges, facilitating the interpretation and practical implementation of the BBN meta-model outputs.
Since these ranges are the maximum of both scales: the meta-models advise can be interpreted as maximizing the PV area and PV power to reach the goal BENG 3 value of $60\%$ to $80\%$.

\subsection{Discussion}
In the end, to validate this result, we finish the loop by calculating the final configuration of the dwelling with the original NTA 8800 model.
The NTA 8800 returns a value of $71\%$, confirming the reverse inference capability of the meta-model. 
These results illustrate in a simple and digestible example how the BBN meta-model is capable of providing valuable insights and assisting architects and engineers in navigating the multidimensional decision space.

By using the numerical Design of Experiments and Sensitivity Analysis we are effectively conducting a dimensionality reduction task similar to the low rank SVD as introduced before. 
As mentioned earlier, the mention of this particular approach of ACD was to illustrate the utility of the framework with a concrete example in a societally relevant context where a Machine Learning approach to modelling can help make an otherwise very complicated simulation procedure to be approximately scaled up massively for policy analysis. 
Here we discuss the potentials and the shortcomings of the model and note the issues with this large-scale black-box approximation that require further investigation. 

The existence of categorical variables in the inputs of the BBN limits the general applicability of ACD as it affects the smoothness and differentiability of the underlying function that is being approximated. 
However, for pragmatic reasons, we have ignored this issue to demonstrate the idea in a large-scale case.  


\paragraph{Validation results}
Compared to the NTA 8800, the Bayesian meta-model is capable of capturing the most important relationships between inputs and outputs. 
However, the difference between the meta-model's predictions and the NTA 8800 predictions can be rather high. 
This means that there is a large difference between the output of the meta-model and the NTA 8800.
The NRMSEs of BENG 1 ($0.82 \%$), and BENG 2 ($1.52 \%$) show that the BBN is able to follow the NTA 8800 to some extent, but is far from accurate as NRMSE is greater than $0.5 \%$. 
On the other hand, the model could predict BENG 3 ($0.47$) relatively accurately. 
This insufficient accuracy was expected as we have dictated a particular topology to the BBN while learning the network structure is an important step in constructing BBNs.
This decision was made to enable the model to function in a bidirectional way: inference and reverse inference (i.e. evaluation and derivation in the terminology of our ACD framework). 

\paragraph{Foreward Inference}
The trained BBN is now capable of inferring the outputs of interest given certain input configurations.
This inference uses the learned CPDs to predict the most likely values for the outputs.
In this way, we can predict the energy performance of buildings, in a quick and intuitive way for ex-ante assesment, based on a certain design configuration. 
In particular, inference demonstrates the potentials of a mapping described in Section \ref{sec:frame}.

\paragraph{Backward Inference}
Since our BBN had only two layers in its network, it can be reversed.
This means that instead of presenting evidence to it, we can present the desired performance values and ask for the derivation of the particular configuration of inputs that will produce such an output.
This can be done through the \emph{Variable Elimination} module of the \emph{pgmpy} \autocite{ankan2015pgmpy}.
The same is also true for a combination of given input-outputs; meaning that evidence can be given for both inputs and outputs of the BBN.
In such cases, the given inputs can also function as the design constraints.
The reverse inference demonstrates how we can utilize probabilistic models to navigate a decision space as explained in Section \ref{sec:frame}.

\paragraph{Augmenting}
The Bayesian meta-model is capable of representing the input-output relationships in a bidirectional and probabilistic format, illustrating a complete example of mapping and navigating processes. 
However, the use of a subset of the most influential variables of the NTA 8800 limits the navigation to decision space made out of the selected variables. 
Nevertheless, this selection was necessary to manage the computational resource-intensive task of learning.
Therefore, BBN does not compete with, or mimic the NTA 8800 model, but rather complements it by increasing its accessibility and providing navigation capabilities. 
The result is a model that can augment the designers' intuition or experience and enhance the level of accuracy even in otherwise vague processes of policy formulation, e.g. in assessing the potential efficacy of alternative subsidies and incentives for building renovation aimed at sustainable energy transition. 

\paragraph{Binning}
The variables are divided into a specified number of bins based on the frequency of the values. In equal frequency binning, the data is first sorted in ascending order and then divided into the specified number of bins, with each bin containing an equal number of observations. Equal frequency binning can be useful for analyzing data that has a skewed distribution or contains outliers. By dividing the data into equal-frequency bins, the impact of outliers may be reduced, and the distribution of the data may be more balanced across the bins. Considering the prediction robustness of the discretized values, each bin should contain a fair amount of data points. A fair amount is a bit vague description because there is no method or rule of thumb for deciding the number of bins. However, we need to keep in mind that the number of bins corresponds with the number of states a parameter can have, and there with the number of parameters that are used to learn the network. Increasing the number of bins, results in an exponential increase of computational demand. Decreasing the number of bins, however, results in too few states to gain the desired insight accuracy in the input-output probabilities. This means that the number of bins should be chosen carefully, keeping in mind the trade-offs, and satisfying the research accuracy and the robustness of the Bayesian meta-model. 

\paragraph{Discrete, Categorical, and Numerical Variables}
The outcome of the inference exercises helped to reveal the relationship between architectural language and engineering behaviour and increases the designer's creative intuition to guide the design process. The design goals of the research were therefore (1) to develop a methodology to gain general and explorative knowledge about the association between spatial and technical building configurations and energy performance in dwellings, and (2) to develop a proposal for an adapted representation of the NTA 8800 model tool, embedded in a computational model, to support intelligent decision making. The research workflow that has been described is a valid approach to reaching the stated design goals.

\paragraph{Toy problem}

The results of the toy problem show the implementation possibilities, and show, of course, simplified, the possibilities for professionals to utilize the tool in the practical field. As shown in the previous chapter, the tool works as expected, and is capable of making preliminary predictions with very few input variables. It appeals to the imagination when the tool is connected to spatial configurations instead of unmeaning variables in a computer script.

\section{Outlook}
\label{sec:outlook}

The ACD framework and its constituent concepts can be best positioned within the context of performance-driven computational design and generative design. 
In particular, the idea of approximating complex and non-linear functions for estimating measurable performance indicators from configurations of decision variables, even if referring to non-spatial decision variables, is generalisable to all areas of computer-aided design. 
However, such surrogate models are not to replace simulation models based on first principles as they can not match their transparency and explainability.
Nevertheless, in cases where one needs to estimate the effects of design decisions on human factors, ergonomics, or combinations of many different types of governing equations, an estimation model trained from actual data can be of utility in that it provides a basis for comparisons in the absence of analytical knowledge.
In other words, the utility of ANNs for \emph{mapping} the associations between decision data points and performance data points is apparent. 

The \emph{navigation} problem, on the other hand, is much harder, philosophically, technically, and mathematically for being solved in any sense.
The real advantage of an AI framework in dealing with a design space navigation problem can be attained if the latent space of the model reveals interpretable information or if it is at least coupled with a sensible low-dimensional space. 
If the latent space of e.g. an Auto Encoder \autocite{marin_spectral_2021} is understandable as a low-dimensional vector space (as an endpoint of the \emph{mapping} and the start point of the \emph{navigating} processes) then it can be used not only to guide the navigation process but also to gain insight as to which design variables are more important in determining the attainment levels of outcomes of interest. 
In other words, even though it appears that in the mapping process, the information content of the decision data points is reducing gradually, one can think of this process as a distillation of an elixir from a large data point that makes the information richer from a human perspective.

In this light, the major advantages of the proposed framework are twofold:
Firstly, providing an elegant framework for applying AI in computational design in the presence of many complex quality criteria; and secondly, providing an elegant framework for designing spatial manifolds very much like the methodology of electrical engineering in designing electronic circuits and systems for signal processing. 
The latter point requires much more space for discussing the theoretical minimum for such an approach to design from a signal processing standpoint. 
In short, however, we can briefly mention that the idea of defining a central representation of a configuration as a discrete manifold provides for directly modelling the functionality of the spatial manifold w.r.t. the flows of walks, light rays, or forces not only from the point of view of spatial movement but also much more elegantly and efficiently in the frequency or spectral domain (which can be attained using Discrete Fourier Transform or Spectral Mesh Analysis). 
One fundamental idea of analogue circuit design from a signal processing point of view is that of designing passive ``filters'' whose properties can much better be understood in the so-called frequency domain analyses put forward by Fourier and Laplace transforms of the so-called transfer functions of the RLC (Resistor, Self-Induction Loop, Capacitor) circuits. 
This approach to circuit design can be traced back to the ideas and propositions of Oliver Heaviside (1850-1925), a self-educated pioneer of electrical engineering. 
Arguably, this frequency-based outlook (relating to the spectrum of eigen frequencies of vibration of shapes, thus also identifiable as a spectral approach), has revolutionized the formation of the field of electronics and thus contributed significantly to the development of AI as we know it today.
Identifying spectral latent spaces and associating them with low-dimensional performance spaces and latent spaces of ANNs is a topic that calls for further theoretical research and computational experimentation.

\AtNextBibliography{\small}
\printbibliography

@misc{regenwetter_towards_2022,
	title = {Towards {Goal}, {Feasibility}, and {Diversity}-{Oriented} {Deep} {Generative} {Models} in {Design}},
	url = {http://arxiv.org/abs/2206.07170},
	urldate = {2022-07-22},
	publisher = {arXiv},
	author = {Regenwetter, Lyle and Ahmed, Faez},
	month = jun,
	year = {2022},
	note = {arXiv:2206.07170 [cs]},
}

@book{pearl_probabilistic_1988,
	title = {Probabilistic reasoning in intelligent systems: networks of plausible inference},
	shorttitle = {Probabilistic reasoning in intelligent systems},
	publisher = {Morgan kaufmann},
	author = {Pearl, Judea},
	year = {1988},
}

@book{koller_probabilistic_2009,
	address = {Cambridge, MA},
	series = {Adaptive computation and machine learning},
	title = {Probabilistic graphical models: principles and techniques},
	isbn = {978-0-262-01319-2},
	shorttitle = {Probabilistic graphical models},
	language = {en},
	publisher = {MIT Press},
	author = {Koller, Daphne and Friedman, Nir},
	year = {2009},
}

@book{friedman_toward_1975,
	address = {Cambridge, Mass},
	edition = {First American Edition},
	title = {Toward a scientific architecture},
	isbn = {978-0-262-06058-5},
	language = {English},
	publisher = {MIT Press},
	author = {Friedman, Yona},
	month = jan,
	year = {1975},
}

@inproceedings{veloso_academy_2020,
	title = {An {Academy} of {Spatial} {Agents}: {Generating} {Spatial} {Configurations} with {Deep} {Reinforcement} {Learning}},
	shorttitle = {An {Academy} of {Spatial} {Agents}},
	author = {Veloso, Pedro and Krishnamurti, Ramesh},
	month = sep,
	year = {2020},
}

@article{batty_theory_1974,
	title = {A {Theory} of {Markovian} {Design} {Machines}},
	doi = {10.1068/b010125},
	journal = {Environment and Planning B: Planning and Design},
	author = {Batty, M.},
	year = {1974},
}

@article{simon_structure_1973,
	title = {The {Structure} of {Ill} {Structured} {P} coblems},
	language = {en},
	journal = {Artificial Intelligence},
	author = {Simon, Herbert A},
	year = {1973},
	pages = {21},
}

@techreport{oshaughnessy_topology_2021,
	type = {preprint},
	title = {Topology {Optimization} using the {Discrete} {Element} {Method}. {Part} 1: {Methodology}, {Validation}, and {Geometric} {Nonlinearity}},
	shorttitle = {Topology {Optimization} using the {Discrete} {Element} {Method}. {Part} 1},
	url = {https://osf.io/c6ymn},
	language = {en},
	urldate = {2021-12-20},
	institution = {engrXiv},
	author = {O'Shaughnessy, Connor and Masoero, Enrico and Gosling, Peter D.},
	month = jul,
	year = {2021},
	doi = {10.31224/osf.io/c6ymn},
}

@book{nourian_configraphics_2016,
	title = {Configraphics: {Graph} {Theoretical} {Methods} for {Design} and {Analysis} of {Spatial} {Configurations}},
	copyright = {Copyright (c) 2016 Pirouz Nourian (Author)},
	isbn = {978-94-6186-720-9},
	url = {https://books.bk.tudelft.nl/press/catalog/book/isbn.9789461867209},
	language = {en},
	urldate = {2021-09-23},
	publisher = {TU Delft Open},
	author = {Nourian, Pirouz},
	month = sep,
	year = {2016},
	doi = {10.7480/isbn.9789461867209},
}

@article{nourian_voxelization_2016,
	title = {Voxelization algorithms for geospatial applications: {Computational} methods for voxelating spatial datasets of {3D} city models containing {3D} surface, curve and point data models},
	volume = {3},
	issn = {2215-0161},
	shorttitle = {Voxelization algorithms for geospatial applications},
	url = {http://www.sciencedirect.com/science/article/pii/S2215016116000029},
	doi = {10.1016/j.mex.2016.01.001},
	language = {en},
	urldate = {2020-04-28},
	journal = {MethodsX},
	author = {Nourian, Pirouz and Gonçalves, Romulo and Zlatanova, Sisi and Ohori, Ken Arroyo and Vu Vo, Anh},
	month = jan,
	year = {2016},
	pages = {69--86},
}

@inproceedings{azadi_godesign_2021,
	address = {Novi Sad, Serbia},
	title = {{GoDesign}: {A} modular generative design framework for mass-customization and optimization in architectural design},
	volume = {1},
	shorttitle = {{GoDesign}},
	booktitle = {Towards a new, configurable architecture},
	publisher = {CUMINCAD},
	author = {Azadi, Shervin and Nourian, Pirouz},
	month = aug,
	year = {2021},
	pages = {285--294},
}

@inproceedings{bai_decision-making_2020,
	title = {Decision-{Making} as a {Social} {Choice} {Game}},
	volume = {2},
	language = {en},
	booktitle = {Proceedings of the 38th {eCAADe} {Conference}},
	author = {Bai, Nan and Azadi, Shervin and Nourian, Pirouz and Roders, Ana Pereira},
	year = {2020},
	pages = {10},
}

@article{regenwetter_deep_2022,
	title = {Deep {Generative} {Models} in {Engineering} {Design}: {A} {Review}},
	volume = {144},
	issn = {1050-0472, 1528-9001},
	shorttitle = {Deep {Generative} {Models} in {Engineering} {Design}},
	url = {https://asmedigitalcollection.asme.org/mechanicaldesign/article/144/7/071704/1136676/Deep-Generative-Models-in-Engineering-Design-A},
	doi = {10.1115/1.4053859},
	language = {en},
	number = {7},
	urldate = {2022-07-24},
	journal = {Journal of Mechanical Design},
	author = {Regenwetter, Lyle and Nobari, Amin Heyrani and Ahmed, Faez},
	month = jul,
	year = {2022},
	pages = {071704},
}

@article{marin_spectral_2021,
	title = {Spectral {Shape} {Recovery} and {Analysis} {Via} {Data}-driven {Connections}},
	volume = {129},
	issn = {0920-5691, 1573-1405},
	url = {https://link.springer.com/10.1007/s11263-021-01492-6},
	doi = {10.1007/s11263-021-01492-6},
	language = {en},
	number = {10},
	urldate = {2022-07-24},
	journal = {Int J Comput Vis},
	author = {Marin, Riccardo and Rampini, Arianna and Castellani, Umberto and Rodolà, Emanuele and Ovsjanikov, Maks and Melzi, Simone},
	month = oct,
	year = {2021},
	pages = {2745--2760},
}

@Inbook{Nourian2023,
author="Nourian, Pirouz
and Azadi, Shervin
and Oval, Robin",
editor="Kyratsis, Panagiotis
and Manavis, Athanasios
and Davim, J. Paulo",
title="Generative Design in Architecture: From Mathematical Optimization to Grammatical Customization",
bookTitle="Computational Design and Digital Manufacturing",
year="2023",
publisher="Springer International Publishing",
address="Cham",
pages="1--43",
isbn="978-3-031-21167-6",
doi="10.1007/978-3-031-21167-6_1"
}

@article{kroes2006dual,
  title={The dual nature of technical artefacts},
  author={Kroes, Peter and Meijers, Anthonie},
  journal={Studies in History and Philosophy of Science},
  volume={37},
  number={1},
  pages={1--4},
  year={2006}
}

@article{kroes2010engineering,
  title={Engineering and the dual nature of technical artefacts},
  author={Kroes, Peter},
  journal={Cambridge journal of economics},
  volume={34},
  number={1},
  pages={51--62},
  year={2010},
  publisher={Oxford University Press}
}

@book{hillier2007space,
  title={Space is the machine: a configurational theory of architecture},
  author={Hillier, Bill},
  year={2007},
  publisher={Space Syntax}
}

@book{giedion2009space,
  title={Space, time and architecture: the growth of a new tradition},
  author={Giedion, Sigfried},
  year={2009},
  publisher={Harvard University Press}
}

@article{Ankenman2010StochasticKrigingFor,
	doi = {10.1287/opre.1090.0754},
	url = {https://doi.org/10.1287%2Fopre.1090.0754},
	year = 2010,
	month = {apr},
	publisher = {Institute for Operations Research and the Management Sciences ({INFORMS})},
	volume = {58},
	number = {2},
	pages = {371--382},
	author = {Bruce Ankenman and Barry L. Nelson and Jeremy Staum},
	title = {Stochastic Kriging for Simulation Metamodeling},
	journal = {Operations Research}
}

@Inbook{conti2021explainable,
  title={Explainable ML: Augmenting the interpretability of numerical simulation using Bayesian networks},
  author={Conti, Zack Xuereb and Kaijima, Sawako},
  booktitle={The Routledge Companion to Artificial Intelligence in Architecture},
  pages={315--335},
  year={2021},
  publisher={Routledge}
}

@inproceedings{BartonSimulationMetamodels,
	doi = {10.1109/wsc.1998.744912},
	url = {https://doi.org/10.1109%2Fwsc.1998.744912},
	publisher = {{IEEE}},
	year={1998},
	author = {R.R. Barton},
	title = {Simulation metamodels},
	booktitle = {1998 Winter Simulation Conference. Proceedings (Cat. No.98CH36274)}
}

@inproceedings{conti2018FlexibleSimulation,
  doi = {10.13140/RG.2.2.23313.53600},
  url = {http://rgdoi.net/10.13140/RG.2.2.23313.53600},
  author = {Conti, Zack Xuereb and Kaijima, Sawako},
  language = {en},
  title = {A Flexible Simulation Metamodel for Exploring Multiple Design Spaces},
  organization={International Association for Shell and Spatial Structures (IASS)},
  year = {2018}
}

@article{Herman2017SalibOpenSource,
	doi = {10.21105/joss.00097},
	url = {https://doi.org/10.21105%2Fjoss.00097},
	year = 2017,
	month = {jan},
	publisher = {The Open Journal},
	volume = {2},
	number = {9},
	pages = {97},
	author = {Jon Herman and Will Usher},
	title = {{SALib}: An open-source Python library for Sensitivity Analysis},
	journal = {The Journal of Open Source Software}
}

@misc{europian_union_2021, 
 	title={Proposal for a Directive of the European Parliament and of the Council on the Energy Performance of Buildings (Recast)}, 
	url={https://eur-lex.europa.eu/legal-content/EN/ALL/?uri=CELEX%3A52021PC0802}, 
	journal={Lex - 52021PC0802 - en - EUR-lex}, 
	author={Europian Union}, 
	year={2021}, 
	month={Dec},
	note = {Accessed on Feb 28th, 2023},
}

@article{Kleijnen1975ACommentBlanning,
	doi = {10.1287/inte.5.3.21},
	url = {https://doi.org/10.1287%2Finte.5.3.21},
	year = {1975},
	month = {may},
	publisher = {Institute for Operations Research and the Management Sciences ({INFORMS})},
	volume = {5},
	number = {3},
	pages = {21--23},
	author = {Jack P. C. Kleijnen},
	title = {A Comment on Blannings Metamodel for Sensitivity Analysis: The Regression Metamodel in Simulation},
	journal = {Interfaces}
}

@misc{gebouwenergieprestatie2022, 
 	title={NTA 8800}, 
	url = {https://www.gebouwenergieprestatie.nl/bepalingsmethode}, 
	journal={Gebouwenergieprestatie (EPG)}, 
	year={2022}, 
	month={Dec},
	note = {Accessed on Feb 28th, 2023}
}

@misc{nen_2017, 
 	title={Nen 7120+C2:2012/A1:2017 NL}, 
	url={https://www.nen.nl/nen-7120-c2-2012-a1-2017-nl-229670}, 
	journal={NEN}, 
	year={2017}, 
	month={Jun},
	note = {Accessed on Feb 28th, 2023},
}

@book{van1991system,
  title={System Design Modeling and Metamodeling},
  author={van Gigch, J.P.},
  isbn={9780306437403},
  lccn={lc91016229},
  series={Language of science},
  url={https://books.google.nl/books?id=M5mD0ZZcwaEC},
  year={1991},
  publisher={Plenum}
}

@Inbook{KleijnenLowOrderPolynomial,
	author={Kleijnen, Jack P.C.},
	title="Low-order polynomial regression metamodels and their designs: basics",
	bookTitle="Design and Analysis of Simulation Experiments",
	year="2008",
	publisher="Springer US",
	address="Boston, MA",
	pages="15--71",
	isbn="978-0-387-71813-2",
	doi="10.1007/978-0-387-71813-2_2"
}

@article{Sacks1989DesignAnalysisComputer,
	doi = {10.1214/ss/1177012413},
	url = {https://doi.org/10.1214%2Fss%2F1177012413},
	year = 1989,
	month = {nov},
	publisher = {Institute of Mathematical Statistics},
	volume = {4},
	number = {4},
	author = {Jerome Sacks and William J. Welch and Toby J. Mitchell and Henry P. Wynn},
	title = {Design and Analysis of Computer Experiments},
	journal = {Statistical Science}
}

@book{Fang2005DesignModelingFor,
	doi = {10.1201/9781420034899},
	url = {https://doi.org/10.1201%2F9781420034899},
	year = 2005,
	month = {oct},
	publisher = {Chapman and Hall/{CRC}},
	author = {Kai-Tai Fang and Runze Li and Agus Sudjianto},
	title = {Design and Modeling for Computer Experiments}
}

@article{sobol1990sensitivity,
  title={On sensitivity estimation for nonlinear mathematical models},
  author={Sobol', Il'ya Meerovich},
  journal={Matematicheskoe modelirovanie},
  volume={2},
  number={1},
  pages={112--118},
  year={1990},
  publisher={Russian Academy of Sciences, Branch of Mathematical Sciences}
}

@inproceedings{ankan2015pgmpy,
  title={pgmpy: Probabilistic graphical models using python},
  author={Ankan, Ankur and Panda, Abinash},
  booktitle={Proceedings of the 14th python in science conference (scipy 2015)},
  volume={10},
  year={2015},
  organization={Citeseer}
}

@article{martin2012extraordinary,
  title={The extraordinary SVD},
  author={Martin, Carla D and Porter, Mason A},
  journal={The American Mathematical Monthly},
  volume={119},
  number={10},
  pages={838--851},
  year={2012},
  publisher={Taylor \& Francis}
}

@inproceedings{soman_aditya_decigenarch_2022,
	title = {{DeciGenArch}: {A} {Generative} {Design} {Methodology} for {Architectural} {Configuration} via {Multi}-{Criteria} {Decision} {Analysis}},
	language = {en},
	booktitle = {Proceedings of {eCAADe} 2022},
	publisher = {Education and research in Computer Aided Architectural Design in Europe},
	author = {{Soman, Aditya} and {Azadi, Shervin} and {Nourian, Pirouz}},
	year = {2022},
	pages = {459-468}
}

@article{weng2021diffusion,
  title   = "What are diffusion models?",
  author  = "Weng, Lilian",
  journal = "lilianweng.github.io",
  year    = "2021",
  month   = "Jul",
  url     = "https://lilianweng.github.io/posts/2021-07-11-diffusion-models/"
}

@incollection{veloso2021mapping,
  title={Mapping generative models for architectural design},
  author={Veloso, Pedro and Krishnamurti, Ramesh},
  booktitle={The Routledge Companion to Artificial Intelligence in Architecture},
  pages={29--58},
  year={2021},
  publisher={Routledge}
}

@article{PENG2010BayesianNetworkReasoning,
	doi = {10.1142/s0218488510006696},
	url = {https://doi.org/10.1142%2Fs0218488510006696},
	year = 2010,
	month = {oct},
	publisher = {World Scientific Pub Co Pte Lt},
	volume = {18},
	number = {05},
	pages = {539--564},
	author = {Yun Peng and Shenyong Zhang and Rong Pan},
	title = {{Bayesian} {Network} {Reasoning} {with} {Uncertain} {Evidences}},
	journal = {International Journal of Uncertainty, Fuzziness and Knowledge-Based Systems}
}

@book{Hicks1964_HICFCI,
	author = {Charles Robert Hicks},
	year = {1964},
	title = {Fundamental Concepts in the Design of Experiments},
	publisher = {New York: Holt, Rinehart and Winston}
}

@book{simon2019sciences,
  title={The Sciences of the Artificial, reissue of the third edition with a new introduction by John Laird},
  author={Simon, Herbert A},
  year={2019},
  publisher={MIT press}
}

@article{gero2004situated,
  title={The situated function--behaviour--structure framework},
  author={Gero, John S and Kannengiesser, Udo},
  journal={Design studies},
  volume={25},
  number={4},
  pages={373--391},
  year={2004},
  publisher={Elsevier}
}

@book{james2013introduction,
  title={An Introduction to Statistical Learning: with Applications in R},
  author={James, G. and Witten, D. and Hastie, T. and Tibshirani, R.},
  isbn={9781461471387},
  lccn={13936251},
  series={Springer Texts in Statistics},
  url={https://books.google.nl/books?id=qcI\_AAAAQBAJ},
  year={2013},
  publisher={Springer New York}
}

@inproceedings{zeng2021musicbert,
author = {Zeng, Mingliang and Tan, Xu and Wang, Rui and Ju, Zeqian and Qin, Tao and Liu, Tie-Yan},
title = {MusicBERT: Symbolic Music Understanding with Large-Scale Pre-Training},
booktitle = {ACL-IJCNLP 2021},
year = {2021},
month = {June},
url = {https://www.microsoft.com/en-us/research/publication/musicbert-symbolic-music-understanding-with-large-scale-pre-training/},
}

@article{jia2023spatial,
title={Spatial decision support systems for hospital layout design: A review},
author={Jia, Zhuoran and Nourian, Pirouz and Luscuere, Peter and Wagenaar, Cor},
journal={Journal of Building Engineering},
pages={106042},
year={2023},
publisher={Elsevier}
}

@article{Bui2021MultiBehaviorWith,
	doi = {10.3390/electronics10091026},
	url = {https://doi.org/10.3390%2Felectronics10091026},
	year = 2021,
	month = {apr},
	publisher = {{MDPI} {AG}},
	volume = {10},
	number = {9},
	pages = {1026},
	author = {Van Bui and Nam Tuan Le and Van Hoa Nguyen and Joongheon Kim and Yeong Min Jang},
	title = {Multi-Behavior with Bottleneck Features {LSTM} for Load Forecasting in Building Energy Management System},
	journal = {Electronics}
}

@article{Khan2021EnsemblePredictionApproach,
	doi = {10.3390/sym13030405},
	url = {https://doi.org/10.3390%2Fsym13030405},
	year = 2021,
	month = {mar},
	publisher = {{MDPI} {AG}},
	volume = {13},
	number = {3},
	pages = {405},
	author = {Anam Nawaz Khan and Naeem Iqbal and Rashid Ahmad and Do-Hyeun Kim},
	title = {Ensemble Prediction Approach Based on Learning to Statistical Model for Efficient Building Energy Consumption Management},
	journal = {Symmetry}
}

\appendix
\pagebreak
\section{Acronyms}

\begin{table}[ht]
     \caption{Acronyms}
     \footnotesize
     \centering
     \setlength\tabcolsep{4pt}
\begin{tabularx}{\hsize}{
    >{\raggedleft\arraybackslash\hsize=0.3\hsize}X
    >{\raggedright\arraybackslash\hsize=1.3\hsize}X
}
\toprule	
\textbf{Acronym}
& \textbf{Term}
\\
\toprule	
ACD
& Augmented Computational Design
\\
AEC
& Architecture, Engineering, and Construction
\\
AI
& Artificial Intelligence
\\
ANN
& Artificial Neural Networks
\\
BBN
& Bayesian Belief Networks
\\
BEM
& Building Energy Modelling
\\
BENG
& Bijna Energie Neutrale Gebouwen: Nearly Zero-Energy Buildings
\\
BIM
& Building Information Model
\\
CAD
& Computer-Aided Design
\\
CEN
& Comité Européen de Normalisation: European Committee of Normalization
\\
CPD
& Conditional Probability Distribution
\\
DAG
& Directed Acyclic Graph
\\
DoE
& Design of Experiment
\\
EPBD
& European Energy Performance of Buildings Directive
\\
JPD
& Joint Probability Distributions
\\
MAPE
& Mean Absolute Percentage Error
\\
MAGMA
& Multi-Attribute Gradient-Driven Mass Aggregation
\\
ML
& Machine Learning
\\
NEN
& Nederlandse Norm: Royal Dutch Standardization Institute
\\
NRMSE
& Normalized Root Mean Square Error
\\
NTA 8800
& Nederlandse Technische Afspraak (Dutch Technical Agreement)
\\
PGM
& Probabilistic Graphical Models
\\
SVD
& Singular Value Decomposition
\\
VAE
& Variational Auto-Encoders
\\

\bottomrule
\end{tabularx}
     \label{tab:nc}
\end{table}
\pagebreak
\section{Notation}

\begin{table}[ht]
     \caption{Notations}
     \footnotesize
     \centering
     \setlength\tabcolsep{4pt}
\begin{tabularx}{\hsize}{
    >{\raggedleft\arraybackslash\hsize=0.5\hsize}X
    >{\raggedright\arraybackslash\hsize=1.1\hsize}X
    >{\raggedright\arraybackslash\hsize=1.4\hsize}X
}
\toprule	
\textbf{Notation}
& \textbf{Name}
& \textbf{Definition}
\\
\toprule	
$\mathbf{x}$
& design/decision space
& $\mathbf{x} \in (0,1]^n$; each $x_i$ corresponds to a single spatial decision variable
\\
$\mathbf{o}$
& performance space
& $\mathbf{o}\in (0,1]^q$; each $o_k$ corresponds to an objective or outcome of interest
\\
$\mathbf{o}=f(\mathbf{x}):=[f_k(\mathbf{x})]_{q\times 1}$
& map from design to performance
& $f: (0,1]^n\mapsto [0,1]^q$; representing a meta-model that approximately maps the decision space to the performance space
\\
$\mathbf{x}=f^{-1}(\mathbf{o})$
& map from performance to design
& $f^{-1}: [0,1]^q \mapsto (0,1]^n$; pseudo-inverse of a meta-model that approximately maps the performance space to the decision space
\\
$\mathbf{J}:=[J_{k,i}]_{q\times n}$
& Jacobian matrix of $f$
& $[J_{k,i}]_{q\times n}=[\frac{\partial f_k}{\partial x_i}]_{q\times n}=[\nabla^T f_q]_{q\times 1}$
\\
$\mathbf{U}_{q\times q}:=[\mathbf{u}_k]_{1\times q}$
& matrix of left singular vectors
& $\mathbf{U}\mathbf{U}^T=\mathbf{U}^T\mathbf{U}=\mathbf{I}_{q\times q}$; ordered by importance
\\
$\mathbf{V}_{n\times n}:=[\mathbf{v}_i]_{1\times n}$
& matrix of right singular vectors
& $\mathbf{V}\mathbf{V}^T=\mathbf{V}^T\mathbf{V}=\mathbf{I}_{n\times n}$; ordered by importance
\\
$\boldsymbol{\Sigma}_{q\times n}:=[\mathbf{u}_k]_{1\times q}$
& matrix of singular value 
& $\boldsymbol{\Sigma}$ is an ${q\times n}$ rectangular diagonal matrix with non-negative real numbers on the diagonal ordered by importance, i.e. singular values $\sigma_c, c\in[0,\min\{q,n\})$
\\

\bottomrule
\end{tabularx}
     \label{tab:notation}
\end{table}

\end{document}